\documentclass{article}
\usepackage{threeparttable}   
\usepackage{diagbox}          
\usepackage{booktabs}
\usepackage{arxiv}
\usepackage{amsmath,amssymb}
\usepackage{algorithm}
\usepackage{algpseudocode}  
\usepackage{subfigure}
\usepackage{algorithm}        
\usepackage{algpseudocode}    
\usepackage[utf8]{inputenc} 
\usepackage[T1]{fontenc}    
\usepackage{hyperref}       
\usepackage{url}            
\usepackage{booktabs}       
\usepackage{amsfonts}       
\usepackage{nicefrac}       
\usepackage{microtype}      
\usepackage{lipsum}         
\usepackage{graphicx}
\usepackage{amsmath}
\usepackage{cleveref}       
\usepackage{multirow}
\usepackage[numbers]{natbib}
\usepackage{graphicx}
\usepackage{doi}

\title{QoSDiff: An Implicit Topological Embedding Learning Framework Leveraging Denoising Diffusion and Adversarial Attention for Robust QoS Prediction}

\newif\ifuniqueAffiliation
\uniqueAffiliationtrue  

\ifuniqueAffiliation
\author{
    \textbf{Guanchen Du} \\
    College of Mathematics and Computer Science\\
    Shantou University, Shantou, Guangdong, China \\
    \texttt{22gcdu@stu.edu.cn} \\
    \And
    \textbf{Jianlong Xu} \thanks{Corresponding author.} \\
    College of Mathematics and Computer Science\\
    Shantou University, Shantou, Guangdong, China \\
    \texttt{xujianlong@stu.edu.cn} \\
    \And
   \textbf{Wei Wei}  \\
College of Mathematics and Computer Science\\
Shantou University, Shantou, Guangdong, China \\
\texttt{22wwei@stu.edu.cn} \\
}
\fi

\date{December 2025}

\renewcommand{\shorttitle}{}

\hypersetup{
pdftitle={QoSDiff: An Implicit Topological Embedding Learning Framework Leveraging Denoising Diffusion and Adversarial Attention for Robust QoS Prediction},
pdfsubject={cs.ML},
pdfauthor={Guanchen Du, Jianlong Xu, Wei Wei},
pdfkeywords={QoS Prediction, Denoising Diffusion Model, Implicit Graph Learning, User-Service Interaction Modeling, Service Computing.},
}

\begin{document}
\maketitle

\begin{abstract}
 Accurate Quality of Service (QoS) prediction is fundamental to service computing, providing essential data-driven guidance for service selection and ensuring superior user experiences. However, prevalent approaches, particularly Graph Neural Networks (GNNs), heavily rely on constructing explicit user--service interaction graphs. Such reliance not only leads to the intractability of explicit graph construction in large-scale scenarios but also limits the modeling of implicit topological relationships and exacerbates susceptibility to environmental noise and outliers. To address these challenges, this paper introduces \emph{QoSDiff}, a novel embedding learning framework that bypasses the prerequisite of explicit graph construction. Specifically, it leverages a denoising diffusion probabilistic model to recover intrinsic latent structures from noisy initializations. To further capture high-order interactions, we propose an adversarial interaction module that integrates a bidirectional hybrid attention mechanism. This adversarial paradigm dynamically distinguishes informative patterns from noise, enabling a dual-perspective modeling of intricate user--service associations. Extensive experiments on two large-scale real-world datasets demonstrate that QoSDiff significantly outperforms state-of-the-art baselines. Notably, the results highlight the framework's superior cross-dataset generalization capability and exceptional robustness against observational noise.
\end{abstract}


\section{Introduction}

Quality of Service (QoS) stands as a crucial non-functional property within the realm of Web services, encompassing essential performance metrics like response time and throughput \cite{zeng2003quality}. Accordingly, QoS prediction is defined as the task of estimating performance-related metrics of service invocations by capturing the underlying interactions between users and services.
With the accelerated development of cloud computing and Internet of Things (IoT) technologies, there has been an explosive increase in Web services that offer comparable functionalities but exhibit varying QoS attributes. Crucially, the absence of robust QoS assessment frameworks frequently results in suboptimal service choices, thereby leading to a marked deterioration in user satisfaction \cite{zou2025privacy}.
It is noteworthy that accurate QoS prediction models are instrumental in supporting the broader service ecosystem. Their predictive outputs are vital for critical downstream applications, including service selection \cite{peng2025energy}, service composition \cite{wu2024constraint}, and service recommendation \cite{cao2024prkg}. Consequently, establishing an effective and dependable QoS prediction methodology has emerged as a core scientific challenge within the domain of service computing \cite{liu2025quality}.

Early research predominantly adopted Collaborative Filtering (CF), which infers missing QoS values by exploiting historical user feedback and modeling similarity structures among users \cite{NCF}. Although CF-based approaches can effectively uncover latent patterns in user–service interactions, they inherently suffer from issues such as the cold-start problem and high computational overhead \cite{zeng2023gatcf}. These drawbacks substantially restrict their scalability and robustness, particularly in dynamic, large-scale, and highly sparse service environments that are characteristic of many real-world applications
\cite{zhang2025survey}.

Since the breakthrough of AlexNet \cite{krizhevsky2012imagenet}, deep learning has demonstrated powerful feature extraction capabilities across a wide range of tasks, particularly in computer vision \cite{xing2025inv}. These capabilities not only significantly improve prediction accuracy but also streamline the overall learning process. Within the QoS prediction domain, Wu et al. \cite{CSMF} proposed CSMF, a model that leverages fully connected neural networks to learn latent representations, thereby enhancing prediction performance. However, fully connected networks are inherently limited in modeling graph-structured data, leading Zeng et al. \cite{zeng2023gatcf} to contend that relying solely on such architectures yields suboptimal results.

With the rise of graph neural networks (GNNs) \cite{GNNs}, significant progress has been made in learning representations for graph-structured data across various domains, including recommender systems and spatial modeling \cite{li2025dual}. Their advantage lies in the ability to incorporate topological information directly into the learning process, enabling more effective embedding of high-order relationships. In the context of QoS prediction, Li et al. \cite{GraphMF} and Liu et al. \cite{QoSGNN} successfully leveraged GNNs to capture the latent structural correlations between users and services. Their methods have shown promising results and established GNNs as a powerful tool in this domain.

Despite the potential shown by graph neural networks in the field of QoS prediction, their practical applications still face three key challenges: 

\begin{enumerate}

    \item \textbf{Intractability of Explicit Graph Construction}: The rapid proliferation of Web services complicates the construction of reliable interaction graphs. In large-scale scenarios, defining explicit edges between massive numbers of users and services not only incurs prohibitive costs in terms of topology modeling but also introduces significant noise due to sparsity. As interaction data expands, maintaining an accurate adjacency structure becomes increasingly intractable for traditional GNNs.
    \item \textbf{Limitations in Modeling Implicit Topological Relationships}: Traditional message-passing-based graph learning methods face considerable challenges in effectively modeling graph structures when explicit topological relationships are absent. This limitation is particularly evident in service invocation scenarios such as those involving cross-domain service combinations, where underlying connections are often implicit.
    \item \textbf{Susceptibility to Environmental Noise and Outliers}: The real-world Web service environment is inherently volatile. QoS data is frequently contaminated by stochastic noise resulting from network fluctuations, server congestion, or temporary failures. Most existing models operate under the assumption of reliable observations and lack dedicated mechanisms to distinguish between intrinsic data patterns and random perturbations. Consequently, their predictive performance deteriorates significantly when exposed to such noisy and unstable environments.
\end{enumerate}

\begin{figure}[]
\centering
\includegraphics[width = 0.5\textwidth]{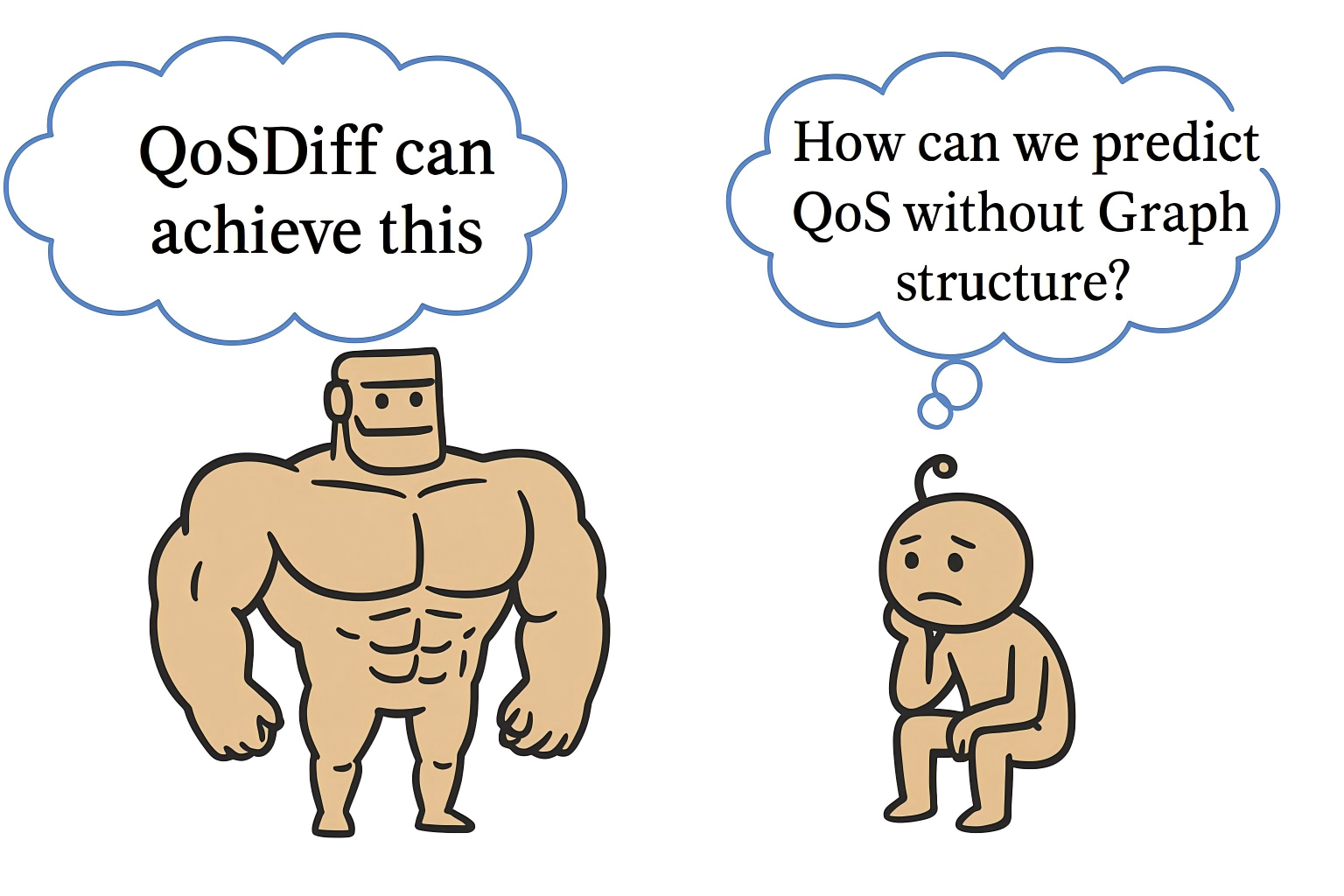}
\caption{\textbf{QoSDiff} can effectively learn embeddings for QoS prediction even in the absence of explicit graph structures.}
\label{fig1}

\end{figure}

To address the aforementioned limitations, we propose \emph{QoSDiff}, a novel QoS prediction framework comprising two core components: a \emph{Diffusion-based Embedding Learning Module} (DELM) and an \emph{Adversarial Attention-based Interaction Module} (AAIM). The DELM is inspired by denoising diffusion probabilistic models (DDPMs) \cite{DDPM} and learns user–service embeddings by progressively denoising latent representations from Gaussian noise. By operating directly on sparse and noisy interaction data rather than relying on fully specified service graphs, DELM alleviates the overhead of explicit graph construction, improves scalability in large-scale service environments, and enhances robustness to missing or perturbed topological information. Unlike conventional DDPMs, which are primarily designed for continuous image data and typically employ UNet-based samplers \cite{unet}, DELM instantiates an attention-based \cite{vaswani2017attention} denoising sampler tailored to the discrete embedding space of QoS observations.

While DELM excels at reconstructing robust static representations from sparse and noisy data, accurate QoS prediction further necessitates capturing the dynamic, reciprocal dependencies inherent in user--service interactions. To bridge the gap between static embedding learning and dynamic interaction modeling, we incorporate the Adversarial Attention-based Interaction Module. 
Unlike traditional CF-based techniques \cite{zeng2023gatcf,QoSGNN} that often implicitly assume a unilateral ``user selects service'' pattern, AAIM adopts a generator--discriminator architecture to progressively refine the interactive representations. Within the generator, we propose a \emph{Bidirectional Hybrid Attention Mechanism} (BHAM) that explicitly models the mutual influence between users and services. This adversarial interplay not only aligns the feature distributions but also serves as a secondary denoising stage, yielding embeddings that are both expressive and resilient to environmental perturbations.

The main contributions of this work are summarized as follows:

\begin{enumerate}
    \item  We propose a DELM that leverages the denoising process of probabilistic diffusion models to learn robust representations directly from interaction data, avoiding explicit graph construction and thereby improving both scalability and robustness to missing or noisy structural information.

    \item Within DELM, we design an attention-driven denoising diffusion sampler specifically tailored to discrete embedding spaces. This sampler performs noise estimation and refinement directly in the discrete domain, overcoming the limitations of continuous UNet-style diffusion architectures.
    
    \item We develop an AAIM equipped with a BHAM, which captures bidirectional user--service dependencies and enhances high-order interaction modeling while mitigating the impact of noisy observations.

    \item We conduct comprehensive experiments on two benchmark QoS datasets, demonstrating clear improvements over existing methods and confirming the contribution of each proposed component through ablation studies.
\end{enumerate}

The rest of this paper is organized as follows.  Section \hyperref[sec:pd]{$\mathrm{II}$} presents the problem definition. In Section \hyperref[sec:method]{$\mathrm{III}$}, we present the technical details of our approach, which is based on the denoising diffusion model framework. Section \hyperref[sec:exp]{$\mathrm{IV}$} provides a detailed presentation of the experiments and their results. Section \hyperref[sec:relatedwork]{$\mathrm{V}$} reviews related work on QoS prediction. Section \hyperref[sec:con]{$\mathrm{VI}$} concludes the paper and discusses future work.

\section{Problem Definitions}
\label{sec:pd}

\begin{figure*}[]
\centering
\includegraphics[width = 0.98\textwidth]{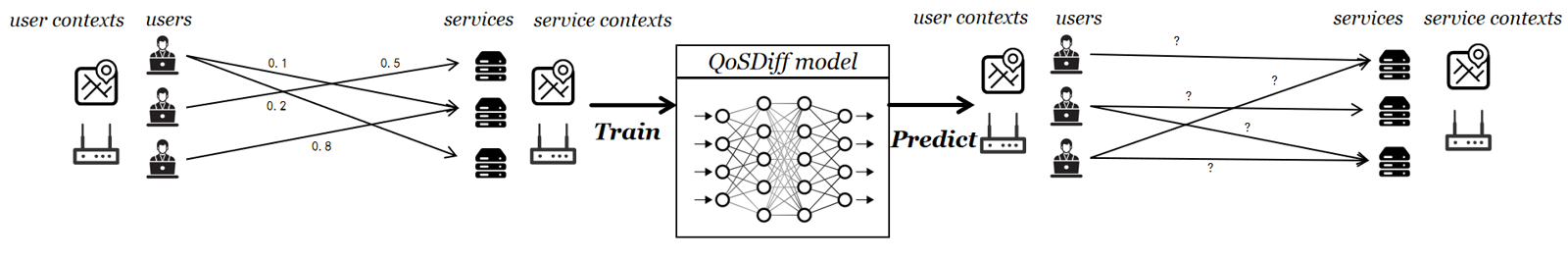}
\caption{Illustration of the QoS prediction task in QoSDiff. 
Observed user--service interactions with known QoS values $r_{ij}$ 
and context attributes are used to train the prediction model $f_\theta$. 
The trained model then estimates the missing QoS values $\hat{r}_{ij}$ 
for unseen user--service pairs $(i,j)\in\bar{\Omega}$.}
\label{fig2}
\end{figure*}

\textbf{Definition 1 (User and Service Sets).} 

Let $U = \{u_1, u_2, \dots, u_m\}$ denote the set of $m$ service users, where each $u_i$ represents an individual requester. Similarly, let $S = \{s_1, s_2, \dots, s_n\}$ denote the set of $n$ accessible Web services, where each $s_j$ represents a unique service provider.

\textbf{Definition 2 (Context Attributes and Embeddings).} 

Let $\mathcal{A}^U = \{A_1^U, A_2^U, \dots, A_p^U\}$ denote the set of available user context fields, where $p$ is the number of attribute types. 
For a specific user $u_i$, let $x_{i}^{(k)}$ represent the raw value/index of the $k$-th attribute in $\mathcal{A}^U$. 
To capture the latent semantics of each attribute, we assign a specific embedding matrix $\mathbf{E}_k^U \in \mathbb{R}^{V_k^U \times d}$ to each context field $k$, where $V_k^U$ is the vocabulary size of that field and $d$ is the embedding dimension. 
Consequently, the raw attribute $y_{j}^{(l)}$ is mapped to a dense vector $\mathbf{e}_{j}^{(k)} \in \mathbb{R}^{d}$. 
Similarly, we define the service context attribute set 
$\mathcal{A}^S = \{A_1^S, A_2^S, \dots, A_q^S\}$ and, for each 
$A_\ell^S$, an embedding matrix $\mathbf{E}_\ell^S \in \mathbb{R}^{V_\ell^S \times d}$,
where $V_\ell^S$ is the vocabulary size of the $\ell$-th service context field.

\textit{Remark:}

The composition of $\mathcal{A}^U$ and $\mathcal{A}^S$ varies across datasets. 
For instance, in the WSDream dataset, the user context set includes $\mathcal{A}^U = \{\text{AS}, \text{Country}, \text{Region}\}$; 
while in the EEL dataset, it focuses on network topology attributes such as $\mathcal{A}^U = \{\text{ISP}, \text{Province}\}$. 
Our framework flexibly adapts to these different context configurations via the aforementioned embedding mechanism.

\textbf{Definition 3 (QoS Matrix).} 

The historical invocation records between users and services are modeled as a user-service QoS matrix $R \in \mathbb{R}^{m \times n}$. Each entry $r_{ij}$ in $R$ represents the observed QoS value (e.g., response time or throughput) of service $s_j$ invoked by user $u_i$. Since a user typically invokes only a limited number of services, 
$R$ is a sparse matrix. Unobserved entries are treated as missing 
values (e.g., denoted by a special symbol “?”) and are not used 
directly as zeros during training.

\textbf{Definition 4 (QoS Prediction Task).}

Given the user set $U = \{u_1, u_2, \dots, u_m\}$, 
the service set $S = \{s_1, s_2, \dots, s_n\}$,
their corresponding context attribute spaces $\mathcal{A}^U$ and $\mathcal{A}^S$,
and the historical QoS matrix $R \in \mathbb{R}^{m \times n}$,
the objective of QoS prediction is to estimate the missing QoS value $r_{ij}$ for any user--service pair $(u_i, s_j)$ where the entry in $R$ is unobserved.

Let $\Omega = \{(i,j) \mid r_{ij} \ \text{is observed}\}$ denote the set of known QoS records, 
and let $\bar{\Omega} = \{(i,j) \mid r_{ij} \ \text{is unobserved}\}$ denote the set of missing entries.
We consider a parametric prediction model $f_\theta$ with learnable parameters $\theta$, and the QoS prediction task aims to learn $\theta$ such that
\[
f_\theta: (u_i, s_j, \mathcal{A}^U, \mathcal{A}^S, R) \rightarrow \hat{r}_{ij},
\]
where the predicted value $\hat{r}_{ij}$ approximates the true QoS value $r_{ij}$ for all $(i,j) \in \bar{\Omega}$.

In other words, the goal is to infer
\[
\hat{r}_{ij} = f_\theta(u_i, s_j, \mathcal{A}^U, \mathcal{A}^S, R), \qquad \forall (i,j) \in \bar{\Omega},
\]
by leveraging (i) the latent embeddings derived from user and service context attributes (as described in Definitions~1--2), 
and (ii) the observed interaction signals encoded in the sparse QoS matrix $R$.

For clarity, we distinguish between raw QoS values and the normalized targets used during training.
Let $\Omega \subseteq \{1,\dots,m\}\times\{1,\dots,n\}$ denote the set of observed user--service pairs.
For each $(i,j)\in\Omega$, we use $r_{ij}$ to denote the original QoS value and $y_{ij}$ to denote its normalized counterpart after applying a fixed preprocessing procedure (e.g., min--max scaling).
The prediction model $f_\theta$ outputs $\hat{y}_{ij}$ as an estimate of $y_{ij}$, and the final predicted QoS $\hat{r}_{ij}$ can be recovered by reversing the normalization if needed.
Unless otherwise specified, all loss functions and evaluation metrics in the following sections are computed on $\{(u_i,s_j,y_{ij})\}$.

\section{Methodology}
\label{sec:method}
\begin{figure*}[h]
    \centering								
    \includegraphics[width = \textwidth]{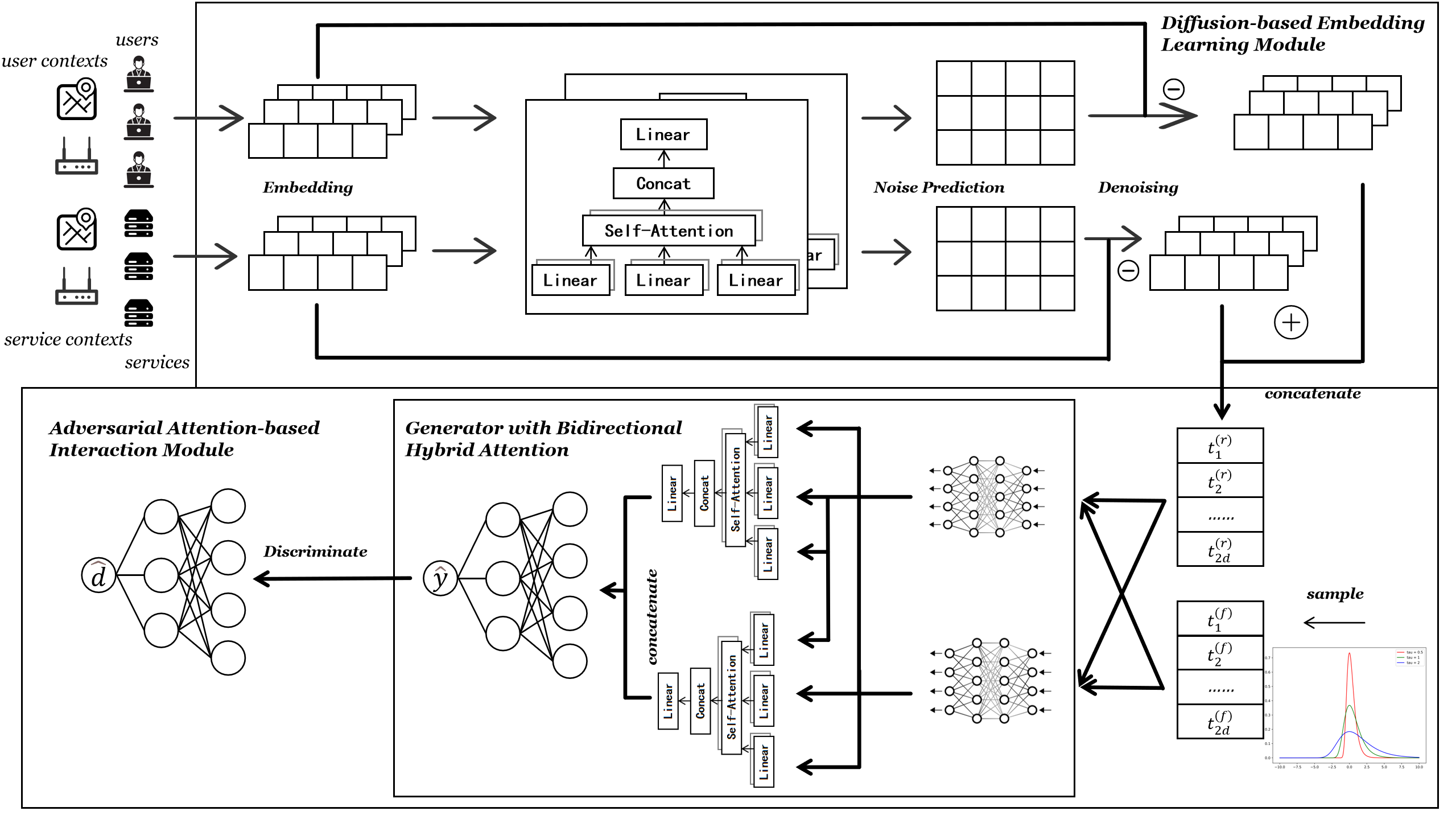}		
    \caption{Overall architecture of the proposed QoSDiff framework. }

    \label{f}
\end{figure*}

In this section, we provide a detailed description of the proposed QoSDiff framework. As outlined in the Introduction, the overall architecture comprises two core components: a Diffusion-based Embedding Learning Module (DELM) and an Adversarial Attention-based Interaction Module (AAIM),  as illustrated in Fig.~\ref{f}. The DELM leverages a denoising diffusion process to learn latent user and service embeddings by progressively denoising representations initialized from a Gaussian distribution. These refined embeddings are subsequently fed into the AAIM, where adversarially trained attention mechanisms capture high-order user–service dependencies, thereby enabling accurate and robust prediction of missing QoS values, even under sparse and noisy observation conditions.

\subsection{Diffusion-based Embedding Learning Module (DELM)}

\subsubsection{Context-aware Embedding Parameterization}
\label{sec:embedding}
Embedding-based parameterization maps discrete user/service identifiers and heterogeneous context fields into a shared continuous latent space, thereby alleviating data sparsity and enabling downstream modules to capture richer semantic relatedness among users and services, while simultaneously providing a well-structured representation space that is amenable to our denoising diffusion–based probabilistic modeling.
Based on Definition~2, we construct context-aware representations for both users
and services by aggregating their identity and attribute-specific embeddings.

On the user side, in addition to the context embedding matrices 
$\{\mathbf{E}_k^U\}_{k=1}^{p}$, we maintain a user identity embedding matrix
$\mathbf{E}_{\mathrm{ID}}^{U} \in \mathbb{R}^{m \times d}$.
For each user $u_i$, the corresponding identity embedding is
\begin{equation}
    \mathbf{e}_i^{\mathrm{ID},U} = \mathbf{E}_{\mathrm{ID}}^{U}[i] \in \mathbb{R}^{d}.
\end{equation}
According to Definition~2, the embedding of the $k$-th user context field
$A_k^{U} \in \mathcal{A}^{U}$ is obtained by indexing the associated embedding
matrix $\mathbf{E}_k^U \in \mathbb{R}^{V_k^U \times d}$ with the raw attribute
value $x_i^{(k)}$, i.e.,
\begin{equation}
    \mathbf{e}_i^{(k),U} = \mathbf{E}_k^U[x_i^{(k)}] \in \mathbb{R}^{d},
    \qquad k = 1,\dots,p.
\end{equation}

Similarly, on the service side we maintain a service identity embedding matrix
$\mathbf{E}_{\mathrm{ID}}^{S} \in \mathbb{R}^{n \times d}$.
For each service $s_j$, its identity embedding is
\begin{equation}
    \mathbf{e}_j^{\mathrm{ID},S} = \mathbf{E}_{\mathrm{ID}}^{S}[j] \in \mathbb{R}^{d}.
\end{equation}
For each service context field $A_\ell^{S} \in \mathcal{A}^{S}$ with embedding
matrix $\mathbf{E}_\ell^S \in \mathbb{R}^{V_\ell^S \times d}$, the corresponding
embedding for $s_j$ is given by
\begin{equation}
    \mathbf{e}_j^{(\ell),S} = \mathbf{E}_\ell^S[y_j^{(\ell)}] \in \mathbb{R}^{d},
    \qquad \ell = 1,\dots,q.
\end{equation}

In this way, both users and services are embedded into a shared $d$-dimensional
latent space, where each representation jointly encodes identity information and
all available context attributes.

\subsubsection{Single-step Denoising Diffusion Formulation}
\label{sec:diffusion}

Instead of relying on GNN-based message passing—which often introduces scalability bottlenecks and over-smoothing issues—we propose a diffusion-guided embedding learning strategy. We conceptualize the embedding initialization process as a reverse denoising generative task. Specifically, we treat the randomly initialized embeddings of users and services as samples drawn from a standard Gaussian distribution (i.e., pure noise). A Denoising Diffusion Probabilistic Model (DDPM) is then employed to iteratively refine these noisy representations into informative latent codes.

Unlike standard DDPMs that require a lengthy Markov chain of $T$ steps, we reframe the embedding learning problem as a \textit{single-step} denoising generative process. We postulate that the randomly initialized embedding matrices are not merely arbitrary noise, but rather can be viewed as the result of a single forward diffusion step applied to an ideal, latent semantic representation. This perspective allows us to efficiently recover the intrinsic structural information by reversing this single noise injection operation.

\paragraph{Forward Process as Initialization}

Let $\mathbf{x}_0 \in \mathbb{R}^d$ denote the ideal, noiseless embedding vector (for a user or service) that perfectly captures the latent semantics. We define a forward diffusion process with a single timestep ($T=1$), which corrupts $\mathbf{x}_0$ into the observed noisy state $\mathbf{x}_1$:
\begin{equation}
    q(\mathbf{x}_1 | \mathbf{x}_0) = \mathcal{N}(\mathbf{x}_1; \sqrt{\alpha_1}\mathbf{x}_0, (1-\alpha_1)\mathbf{I}),
\end{equation}
where $\alpha_1 = 1 - \beta_1$ controls the signal-to-noise ratio.
In our framework, the observed state $\mathbf{x}_1$ corresponds to the actual initialized values in our embedding lookup tables. To theoretically align our diffusion formulation with standard deep learning initialization practices, we calibrate the noise schedule based on the properties of Kaiming Initialization. Specifically, for an embedding dimension $d$, Kaiming initialization samples weights from $\mathcal{N}(0, \frac{2}{d}\mathbf{I})$. By equating the variance of the diffusion noise term to this initialization variance, we explicitly set the noise schedule parameters as:
\begin{equation}
    \beta_1 = \frac{2}{d}, \quad \alpha_1 = 1 - \frac{2}{d}.
\end{equation}
This alignment allows us to approximately interpret the initialized embeddings as
\begin{equation}
    \mathbf{x}_1 = \sqrt{\alpha_1}\mathbf{x}_0 + \sqrt{1-\alpha_1}\boldsymbol{\epsilon}, \quad \boldsymbol{\epsilon} \sim \mathcal{N}(\mathbf{0}, \mathbf{I}).
\end{equation}

\paragraph{Single-Step Reconstruction}
Our goal is to reverse this process to recover the ideal embedding $\mathbf{x}_0$. We employ a parameterized denoising network $\boldsymbol{\epsilon}_\theta$ (implemented via the Attention mechanism described in Sec.~\ref{sec:embedding}) to estimate the noise component $\boldsymbol{\epsilon}$ present in $\mathbf{x}_1$.
Based on the DDPM reverse sampling derivation, the reconstruction of the ideal embedding $\hat{\mathbf{x}}_0$ is formulated as:
\begin{equation}
    \hat{\mathbf{x}}_0 = \frac{1}{\sqrt{\alpha_1}} \left( \mathbf{x}_1 - \sqrt{1-\alpha_1} \boldsymbol{\epsilon}_\theta(\mathbf{x}_1) \right) + \sqrt{\beta_1}\mathbf{z},
\end{equation}
where $\mathbf{z} \sim \mathcal{N}(\mathbf{0}, \mathbf{I})$ represents the stochasticity in the generative process.
Substituting the derived Kaiming parameters $\alpha_1 = 1 - \frac{2}{d}$, the final update rule for refining any arbitrary embedding vector $\mathbf{e}$ becomes:
\begin{equation}
\label{eq:refine_rule}
    \hat{\mathbf{e}} = \frac{1}{\sqrt{1 - \frac{2}{d}}} \left( \mathbf{e} - \sqrt{\frac{2}{d}} \boldsymbol{\epsilon}_\theta(\mathbf{e}) \right) + \sqrt{\frac{2}{d}}\mathbf{z}.
\end{equation}
This formulation explicitly subtracts the predicted network noise from the initialized matrix to approach the ideal low-noise state.

\subsubsection{Attention-based Noise Prediction Network} 
While standard DDPMs rely on CNN-based U-Net\cite{unet} architectures, such designs are ill-suited for our task due to their dependence on spatial locality—an inductive bias intrinsic to images but absent in discrete identity embeddings. Applying convolution kernels to non-grid latent vectors imposes artificial structural constraints. 

To address this, we propose a lightweight Attention-based Noise Prediction Module. Unlike CNNs, the self-attention mechanism $\boldsymbol{\epsilon}_\theta$ naturally captures global dependencies across latent dimensions without spatial assumptions, offering a more structurally aligned solution for vector-based denoising. 

\begin{equation} 
\boldsymbol{\epsilon}_\theta(\mathbf{x}) = \mathrm{Linear}(\mathrm{MultiHeadAttn}(\mathbf{x}, \mathbf{x}, \mathbf{x}), 
\end{equation} 
where the Query, Key, and Value are all derived from the current state $\mathbf{x}$. This design allows the model to recalibrate the feature dimensions during the denoising process dynamically.

\subsubsection{Parallel Attribute Refinement and Aggregation}

To capture fine-grained semantic features, we apply the denoising rule in Eq.~(\ref{eq:refine_rule}) \textit{independently} to each identity and context attribute embedding before aggregation.
Specifically, for the generic variable $\mathbf{e}$ in Eq.~(\ref{eq:refine_rule}), we substitute it with the specific embedding components defined in Sec.~\ref{sec:embedding}:
\begin{itemize}
    \item \textbf{User Side:} The user identity embedding $\mathbf{e}_i^{\mathrm{ID},U}$ and each user context attribute embedding $\mathbf{e}_i^{(k),U}$ (where $k=1 \dots p$, e.g., AS, Country).
    \item \textbf{Service Side:} The service identity embedding $\mathbf{e}_j^{\mathrm{ID},S}$ and each service context attribute embedding $\mathbf{e}_j^{(\ell),S}$ (where $\ell=1 \dots q$, e.g., AS, Provider).
\end{itemize}
Distinct attention-based denoising networks are trained for each attribute type to capture their specific data distributions.
After obtaining the refined embeddings (denoted as $\hat{\mathbf{e}}$), we aggregate them to form the final holistic representations.
For a user $u_i$, the final latent vector $\mathbf{z}_i^{U}$ is obtained by summing the refined components followed by layer normalization:
\begin{equation}
    \mathbf{z}_i^{U} = \mathrm{LayerNorm}\left( \hat{\mathbf{e}}_i^{\mathrm{ID},U} + \sum_{k=1}^{p} \hat{\mathbf{e}}_i^{(k),U} \right).
\end{equation}
Similarly, for a service $s_j$, the final representation $\mathbf{z}_j^{S}$ is derived as:
\begin{equation}
    \mathbf{z}_j^{S} = \mathrm{LayerNorm}\left( \hat{\mathbf{e}}_j^{\mathrm{ID},S} + \sum_{\ell=1}^{q} \hat{\mathbf{e}}_j^{(\ell),S} \right).
\end{equation}
This separate-then-aggregate strategy ensures that the intrinsic structure of each attribute is robustly recovered before they are fused for downstream interaction modeling.

The overall procedure of the Diffusion-based Embedding Learning Module (DELM) is summarized in Algorithm~\ref{alg:delm}.
\begin{algorithm}[t]
\caption{Diffusion-based Embedding Learning Module (DELM)}
\label{alg:delm}
\begin{algorithmic}[1]
\Require
    User set $U = \{u_1,\dots,u_m\}$, service set $S = \{s_1,\dots,s_n\}$;
    user context values $\{x_i^{(k)}\}$ for $k=1,\dots,p$;
    service context values $\{y_j^{(\ell)}\}$ for $\ell=1,\dots,q$;
    embedding matrices $\mathbf{E}_{\mathrm{ID}}^{U}$, $\{\mathbf{E}_k^U\}_{k=1}^{p}$,
    $\mathbf{E}_{\mathrm{ID}}^{S}$, $\{\mathbf{E}_\ell^S\}_{\ell=1}^{q}$;
    diffusion parameters $\alpha_1, \beta_1$ (e.g., $\beta_1 = 2/d$, $\alpha_1 = 1 - 2/d$).
\Ensure
    Refined user embeddings $\{\mathbf{z}_i^{U}\}_{i=1}^{m}$ and service embeddings $\{\mathbf{z}_j^{S}\}_{j=1}^{n}$.
\\

\Comment{Initialization as a single-step forward diffusion}
\State Initialize all embedding matrices by Kaiming initialization
       so that each row $\sim \mathcal{N}(\mathbf{0}, \tfrac{2}{d}\mathbf{I})$.

\\
\Procedure{SingleStepReconstruct}{$\mathbf{e}, \boldsymbol{\epsilon}_\theta, \alpha_1, \beta_1$}
    \State Sample $\mathbf{z} \sim \mathcal{N}(\mathbf{0}, \mathbf{I})$
    \State $\hat{\boldsymbol{\epsilon}} \gets \boldsymbol{\epsilon}_\theta(\mathbf{e})$
    \State $\hat{\mathbf{e}} \gets
        \dfrac{1}{\sqrt{\alpha_1}}\big(\mathbf{e} - \sqrt{1-\alpha_1}\,\hat{\boldsymbol{\epsilon}}\big)
        + \sqrt{\beta_1}\,\mathbf{z}$
    \State \Return $\hat{\mathbf{e}}$
\EndProcedure

\\
\Comment{User-side refinement}
\For{each user $u_i \in U$}
    \State $\mathbf{e}_i^{\mathrm{ID},U} \gets \mathbf{E}_{\mathrm{ID}}^{U}[i]$
    \For{$k = 1$ to $p$}
        \State $\mathbf{e}_i^{(k),U} \gets \mathbf{E}_k^U[x_i^{(k)}]$
    \EndFor
    \State $\hat{\mathbf{e}}_i^{\mathrm{ID},U}
        \gets \Call{SingleStepReconstruct}{\mathbf{e}_i^{\mathrm{ID},U},
        \boldsymbol{\epsilon}_\theta^{\mathrm{ID},U}, \alpha_1, \beta_1}$
    \For{$k = 1$ to $p$}
        \State $\hat{\mathbf{e}}_i^{(k),U}
            \gets \Call{SingleStepReconstruct}{\mathbf{e}_i^{(k),U},
            \boldsymbol{\epsilon}_\theta^{(k),U}, \alpha_1, \beta_1}$
    \EndFor
    \State $\mathbf{z}_i^{U} \gets \mathrm{LayerNorm}
        \big(\hat{\mathbf{e}}_i^{\mathrm{ID},U}
        + \sum_{k=1}^{p} \hat{\mathbf{e}}_i^{(k),U} \big)$
\EndFor

\\
\Comment{Service-side refinement}
\For{each service $s_j \in S$}
    \State $\mathbf{e}_j^{\mathrm{ID},S} \gets \mathbf{E}_{\mathrm{ID}}^{S}[j]$
    \For{$\ell = 1$ to $q$}
        \State $\mathbf{e}_j^{(\ell),S} \gets \mathbf{E}_\ell^S[y_j^{(\ell)}]$
    \EndFor
    \State $\hat{\mathbf{e}}_j^{\mathrm{ID},S}
        \gets \Call{SingleStepReconstruct}{\mathbf{e}_j^{\mathrm{ID},S},
        \boldsymbol{\epsilon}_\theta^{\mathrm{ID},S}, \alpha_1, \beta_1}$
    \For{$\ell = 1$ to $q$}
        \State $\hat{\mathbf{e}}_j^{(\ell),S}
            \gets \Call{SingleStepReconstruct}{\mathbf{e}_j^{(\ell),S},
            \boldsymbol{\epsilon}_\theta^{(\ell),S}, \alpha_1, \beta_1}$
    \EndFor
    \State $\mathbf{z}_j^{S} \gets \mathrm{LayerNorm}
        \big(\hat{\mathbf{e}}_j^{\mathrm{ID},S}
        + \sum_{\ell=1}^{q} \hat{\mathbf{e}}_j^{(\ell),S} \big)$
\EndFor

\end{algorithmic}
\end{algorithm}

\textbf{Framework Analysis: Generative Embedding Learning vs. Graph Neural Networks}

The proposed framework shifts the paradigm from predictive graph learning to generative denoising, effectively resolving three inherent limitations of GNN-based methods:

Addressing the intractability of explicit graph construction, our approach eliminates the dependency on pre-defined adjacency matrices. While traditional GNNs face quadratic computational costs and noise from sparse edge definition, our DDPM-based model operates directly within the continuous latent space. By conceptualizing embedding learning as a reverse diffusion process, we bypass massive graph materialization, ensuring scalability and immunity to topology-induced noise in large-scale service ecosystems.

To overcome GNNs' limitation in modeling implicit relationships—typically restricted by local message passing and the homophily assumption—we employ a global attention-based noise estimator. This mechanism captures all-to-all dependencies, identifying latent semantic alignments between topologically distant users and services. Moreover, by integrating global graph statistics during the diffusion phase, our framework generates informative priors for low-degree nodes, significantly enhancing representation quality in sparse and cold-start settings~\cite{hevapathige2025beyond}.

 Regarding susceptibility to environmental noise, our framework offers superior stability compared to deterministic GNNs, which often amplify stochastic outliers. Grounded in probabilistic diffusion theory, our objective is explicitly designed to recover intrinsic signals from Gaussian noise. This inherent "denoising" mechanism acts as a robust filter against network fluctuations and data outliers, yielding stable embeddings even in volatile environments.

\subsection{Adversarial Attention-based Interaction Module (AAIM)}

Built upon the diffusion-based embeddings learned in the previous subsection, we now turn to modeling the high-order interactions between users and services. Classical MF–based \cite{PMF} approaches project users and services into a shared latent space and estimate unknown QoS values via inner products. While effective at capturing simple linear associations, such MF formulations fall short in modeling the complex nonlinear dependencies that underlie real user–service interactions, and they are further hampered by sparsity and noise in practical Web service environments. Likewise, CF-based \cite{zeng2023gatcf} methods exploit user–user or service–service similarity for recommendation, but their performance often degrades severely in sparse or cold-start scenarios, and they are highly sensitive to noisy observations.

To address these limitations, we propose an Adversarial Attention-based Interaction Module (AAIM) that couples a generative adversarial network with a bidirectional hybrid attention mechanism. In our design, the generator adopts a dual-perspective attention architecture that simultaneously models users’ attention over services and services’ attention over users, thereby capturing mutual, high-order user–service dependencies in a unified framework. The discriminator, in turn, is trained to differentiate real interaction patterns—constructed from the learned user and service embeddings—from synthetic samples generated via Gumbel–Softmax–based perturbations. This adversarial training paradigm encourages the generator to produce interaction representations that are both expressive and robust to noise, ultimately leading to more accurate QoS prediction in sparse and complex service environments.

\subsubsection{Real and Synthetic Interaction Embeddings}

Specifically, after the embedding learning stage, we obtain refined user-side and service-side representations, denoted by $\mathbf{z}_i^{U}$ and $\mathbf{z}_j^{S}$, respectively. To better model the complex high-order dependencies encoded in these representations, we first concatenate the two matrices to construct a unified interaction matrix $\mathbf{T}$, formulated as follows:

\begin{equation}
    \label{eq5.1.16}
    \mathbf{T} = \mathbf{z}_i^{U} \,\|\, \mathbf{z}_j^{S},
\end{equation}
where $\mathbf{T}$ denotes the unified interaction representation obtained by concatenating the user-side embedding $\mathbf{z}_i^{U}$ and the service-side embedding $\mathbf{z}_j^{S}$. This matrix serves as the ``real'' input embedding to the prediction layer.

To enable adversarial training, we further construct a synthetic (fake) counterpart. Unlike standard GANs that generate samples from scratch, we employ a \emph{Noise Injection Mechanism} to create synthetic interaction patterns. This approach ensures that the fake embeddings share the same continuous support space as the real embeddings but lack the intrinsic semantic structure, forcing the discriminator to learn meaningful feature distributions.

To generate the fake interaction embedding, we first sample a noise matrix from a standard Gaussian distribution:
\begin{equation}
    \label{eq:noise_input}
    \mathbf{N} \sim \mathcal{N}(0, 1) \in \mathbb{R}^{B \times 2d},
\end{equation}
where $B$ is the batch size and $2d$ corresponds to the concatenated dimension. 

To introduce further stochasticity and cover a broader noise distribution, we apply a scaled perturbation:
\begin{equation}
    \label{eq:fake_sample}
    \mathbf{F} = \mathbf{N} + \tau \cdot \boldsymbol{\epsilon}, \quad \boldsymbol{\epsilon} \sim \mathcal{U}(-1, 1),
\end{equation}
where $\boldsymbol{\epsilon}$ represents uniform noise and $\tau$ is a scaling factor controlling the noise intensity. This resulting matrix $\mathbf{F}$ serves as the fake input embedding. By training the discriminator to distinguish the structured $\mathbf{T}$ from the unstructured $\mathbf{F}$, the generator is incentivized to produce highly robust interaction representations that are resilient to random perturbations.

\subsubsection{Detail of Generator(Bidirectional Hybrid Attention Mechanism )}

Given either a real interaction embedding or a Gumbel--Softmax--generated pseudo-sample, the generator is responsible for producing a refined interaction representation and the corresponding QoS prediction. 
Let $\mathbf{z}_i^{U} \in \mathbb{R}^{d}$ and $\mathbf{z}_j^{S} \in \mathbb{R}^{d}$ denote the refined user and service embeddings obtained from the previous embedding learning stage. 
For a mini-batch of $B$ user–service pairs $(u_i, s_j)$, we first construct the batch-wise interaction matrix
\begin{equation}
    \mathbf{T} = 
    \big[
        \mathbf{t}_{ij}^{(1)}; \mathbf{t}_{ij}^{(2)}; \dots; \mathbf{t}_{ij}^{(B)}
    \big] 
    \in \mathbb{R}^{B \times 2d},
\end{equation}
where each row corresponds to a concatenated interaction vector
\begin{equation}
    \mathbf{t}_{ij}^{(b)} = \mathbf{z}_i^{U} \,\|\, \mathbf{z}_j^{S} \in \mathbb{R}^{2d}.
\end{equation}
In the real-sample branch, $\mathbf{T}$ is instantiated by the unified interaction embedding in \eqref{eq5.1.16}, whereas in the fake-sample branch it is replaced by the Gumbel--Softmax pseudo-sample $\mathbf{F}$ constructed in the previous subsection.

To capture directional dependencies from users to services and from services to users, the generator instantiates a bidirectional hybrid attention block. 
Concretely, we first obtain two projected interaction representations through distinct linear transformations followed by a nonlinearity:
\begin{equation}
    \mathbf{H}^{(U \rightarrow S)} = \sigma\big(\mathbf{T}\mathbf{W}_1^{\top} + \mathbf{1}\mathbf{b}_1^{\top}\big), 
\end{equation}
\begin{equation}
    \mathbf{H}^{(S \rightarrow U)} = \sigma\big(\mathbf{T}\mathbf{W}_2^{\top} + \mathbf{1}\mathbf{b}_2^{\top}\big),
\end{equation}
where $\sigma(\cdot)$ is the ReLU activation, $\mathbf{W}_1, \mathbf{W}_2 \in \mathbb{R}^{d_h \times 2d}$ and $\mathbf{b}_1, \mathbf{b}_2 \in \mathbb{R}^{d_h}$ are learnable parameters, $d_h$ denotes the hidden dimensionality and $\mathbf{1} \in \mathbb{R}^{B}$ is an all-ones column vector. 
The $b$-th rows of $\mathbf{H}^{(U \rightarrow S)}$ and $\mathbf{H}^{(S \rightarrow U)}$,
denoted by $\mathbf{h}^{(U \rightarrow S)}_{ij}$ and $\mathbf{h}^{(S \rightarrow U)}_{ij}$, are then treated as one-step sequences and fed into two multi-head attention (MHA) blocks to model mutual interactions:
\begin{align}
    \tilde{\mathbf{h}}^{(U \rightarrow S)}_{ij} 
        &= \operatorname{MHA}_{U \rightarrow S}\big(\mathbf{h}^{(U \rightarrow S)}_{ij}, \mathbf{h}^{(U \rightarrow S)}_{ij}, \mathbf{h}^{(S \rightarrow U)}_{ij}\big), \\
    \tilde{\mathbf{h}}^{(S \rightarrow U)}_{ij} 
        &= \operatorname{MHA}_{S \rightarrow U}\big(\mathbf{h}^{(S \rightarrow U)}_{ij}, \mathbf{h}^{(S \rightarrow U)}_{ij}, \mathbf{h}^{(U \rightarrow S)}_{ij}\big),
\end{align}
where $\operatorname{MHA}_{U \rightarrow S}(\cdot)$ and $\operatorname{MHA}_{S \rightarrow U}(\cdot)$ denote two independent multi-head attention modules with shared embedding size $d_h$ and a predefined number of heads. 
The first MHA block emphasizes how a user attends to different aspects of the service representation, while the second focuses on how the service attends back to the user, thus forming a bidirectional interaction pattern.

The outputs of the two attention branches are concatenated and passed through a lightweight feed-forward network with layer normalization:
\begin{align}
    \mathbf{g}_{ij} &= \tilde{\mathbf{h}}^{(U \rightarrow S)}_{ij} \,\|\, \tilde{\mathbf{h}}^{(S \rightarrow U)}_{ij} \in \mathbb{R}^{2d_h}, \\
    \mathbf{g}^{(1)}_{ij} &= \operatorname{LN}_1\!\big(\sigma(\mathbf{W}_3 \mathbf{g}_{ij} + \mathbf{b}_3)\big), \\
    \mathbf{g}^{(2)}_{ij} &= \operatorname{LN}_2\!\big(\sigma(\mathbf{W}_4 \mathbf{g}^{(1)}_{ij} + \mathbf{b}_4)\big),
\end{align}
where $\mathbf{W}_3 \in \mathbb{R}^{d_g \times 2d_h}$, $\mathbf{W}_4 \in \mathbb{R}^{d_o \times d_g}$, $\mathbf{b}_3 \in \mathbb{R}^{d_g}$, $\mathbf{b}_4 \in \mathbb{R}^{d_o}$ are learnable parameters, $d_g$ and $d_o$ denote intermediate dimensions, and $\operatorname{LN}_1(\cdot)$, $\operatorname{LN}_2(\cdot)$ are layer normalization operators.

Finally, the generator produces a scalar prediction through a sigmoid-activated output layer:
\begin{equation}
    \hat{y}_{ij} = \sigma\big(\mathbf{w}_5^{\top} \mathbf{g}^{(2)}_{ij} + b_5\big),
\end{equation}
where $\mathbf{w}_5 \in \mathbb{R}^{d_o}$ and $b_5 \in \mathbb{R}$ are trainable parameters. 
The resulting $\hat{y}_{ij} \in (0, 1)$ represents either the predicted QoS value (after appropriate rescaling) or, in the adversarial setting, the generator's confidence score associated with the given interaction input. 
By jointly modeling user-to-service and service-to-user attention within a unified architecture, the generator can capture rich, high-order user–service dependencies and produce expressive interaction representations for subsequent adversarial optimization.

\subsubsection{Detail of Discriminator}

The discriminator is designed to distinguish real interaction signals from those synthesized by the generator, and thereby provide an adversarial supervision signal for training the whole AAIM framework. 
Given a scalar interaction score $s_{ij}$ associated with a user–service pair $(u_i, s_j)$ (e.g., a ground-truth QoS value or the generator's predicted score $\hat{y}_{ij}$), the discriminator maps $s_{ij}$ to a real-valued credibility score $D(s_{ij})$ that reflects how likely the input originates from the real data distribution.

Formally, we first normalize the input into a one-dimensional vector and pass it through a lightweight multi-layer perceptron with leaky-ReLU activations, batch normalization, and dropout regularization:
\begin{align}
    \mathbf{h}^{(1)}_{ij} 
        &= \operatorname{LReLU}\big(\mathbf{W}^{(1)} s_{ij} + \mathbf{b}^{(1)}\big), \\
    \tilde{\mathbf{h}}^{(1)}_{ij} 
        &= \operatorname{Dropout}\!\left(\operatorname{BN}^{(1)}\!\left(\mathbf{h}^{(1)}_{ij}\right)\right), \\
    \mathbf{h}^{(2)}_{ij} 
        &= \operatorname{LReLU}\big(\mathbf{W}^{(2)} \tilde{\mathbf{h}}^{(1)}_{ij} + \mathbf{b}^{(2)}\big), \\
    \tilde{\mathbf{h}}^{(2)}_{ij} 
        &= \operatorname{Dropout}\!\left(\operatorname{BN}^{(2)}\!\left(\mathbf{h}^{(2)}_{ij}\right)\right),
\end{align}
where $\mathbf{W}^{(1)}, \mathbf{W}^{(2)}$ and $\mathbf{b}^{(1)}, \mathbf{b}^{(2)}$ are learnable weight matrices and bias vectors, $\operatorname{LReLU}(\cdot)$ denotes the leaky-ReLU activation, $\operatorname{BN}^{(1)}(\cdot)$ and $\operatorname{BN}^{(2)}(\cdot)$ are batch normalization layers, and $\operatorname{Dropout}(\cdot)$ is a dropout operator with predefined keep probabilities. 
The hidden dimensionality of $\mathbf{h}^{(1)}_{ij}$ and $\mathbf{h}^{(2)}_{ij}$ is denoted by $d_D$, which controls the capacity of the discriminator.

The final discrimination score is obtained through a sigmoid-activated output neuron with a fixed scaling factor:
\begin{equation}
    D(s_{ij}) = \gamma \cdot \sigma\big(\mathbf{w}^{\top} \tilde{\mathbf{h}}^{(2)}_{ij} + b^{(3)}\big),
\end{equation}
where $\mathbf{w} \in \mathbb{R}^{d_D}$ and $b^{(3)} \in \mathbb{R}$ are trainable parameters, $\sigma(\cdot)$ denotes the logistic sigmoid function, and $\gamma > 0$ is a constant scaling coefficient that stretches the output range for numerical stability.

During adversarial training, the discriminator is optimized to assign higher scores $D(s_{ij})$ to real interaction samples (constructed from ground-truth QoS observations and their corresponding embeddings) and lower scores to fake samples produced by the generator. 
Conversely, the generator is trained to produce synthetic interaction scores that maximize $D(s_{ij})$, thus encouraging the learned interaction representations to be indistinguishable from those derived from real data and enhancing the robustness of QoS prediction under sparse and noisy conditions.

\subsubsection{Forward Propagation of AAIM}

To summarize the above components, we now describe the forward propagation of the proposed AAIM module for a given batch of user--service pairs and the procedure is summarized in \ref{alg:aai_forward}. 
For each observed interaction $(u_i, s_j) \in \mathcal{O}$, we first construct the real interaction embedding $\mathbf{t}^{(r)}_{ij}$ and the synthetic (fake) embedding $\mathbf{t}^{(f)}_{ij}$:
\begin{equation}
    \mathbf{t}^{(r)}_{ij} = \mathbf{z}_i^{U} \,\|\, \mathbf{z}_j^{S}, 
    \qquad
    \mathbf{t}^{(f)}_{ij} = \mathbf{F}_{ij},
\end{equation}
where $\mathbf{F}_{ij}$ denotes the Gumbel--Softmax--based pseudo-sample introduced in the previous subsection.

Both embeddings are then fed into the generator to produce two predicted QoS scores:
\begin{equation}
    \hat{y}^{(r)}_{ij} = G\!\big(\mathbf{t}^{(r)}_{ij}\big), 
    \qquad
    \hat{y}^{(f)}_{ij} = G\!\big(\mathbf{t}^{(f)}_{ij}\big),
\end{equation}
where $G(\cdot)$ denotes the bidirectional hybrid attention generator described above. 
The corresponding discriminator outputs are obtained by
\begin{equation}
    \hat{d}^{(r)}_{ij} = D\!\big(\hat{y}^{(r)}_{ij}\big), 
    \qquad
    \hat{d}^{(f)}_{ij} = D\!\big(\hat{y}^{(f)}_{ij}\big),
\end{equation}
where $D(\cdot)$ is the discriminator defined in the previous subsection.

In other words, for each interaction, we obtain four key quantities:
the generator predictions on real and fake embeddings, $\hat{y}^{(r)}_{ij}$ and $\hat{y}^{(f)}_{ij}$, and the corresponding discriminator scores, $\hat{d}^{(r)}_{ij}$ and $\hat{d}^{(f)}_{ij}$. 
Together with the ground-truth QoS value $y_{ij}$, these variables form the basis for our training objectives, which are detailed in the next subsection.

\begin{algorithm}[t]
\caption{Forward Propagation of AAIM}
\label{alg:aai_forward}
\begin{algorithmic}[1]
\Require
    Batch of observed user--service pairs $\mathcal{B} \subseteq \mathcal{O}$;
    refined embeddings $\{\mathbf{z}_i^{U}\}$, $\{\mathbf{z}_j^{S}\}$;
    noise scale $\tau$; generator $G(\cdot)$; discriminator $D(\cdot)$
\Ensure
    Real and fake predictions $\hat{\mathbf{y}}^{(r)}, \hat{\mathbf{y}}^{(f)}$;
    discriminator scores $\hat{\mathbf{d}}^{(r)}, \hat{\mathbf{d}}^{(f)}$
\\[0.5ex]

\State Initialize empty vectors
    $\hat{\mathbf{y}}^{(r)}, \hat{\mathbf{y}}^{(f)},
     \hat{\mathbf{d}}^{(r)}, \hat{\mathbf{d}}^{(f)}$
\State Compute batch size $B \gets |\mathcal{B}|$

\Comment{Construct real interaction embeddings}
\State For each $(u_i, s_j) \in \mathcal{B}$:
\State \hspace{1.5em} $\mathbf{t}^{(r)}_{ij}
    \gets \mathbf{z}_i^{U} \,\|\, \mathbf{z}_j^{S}$
\State Stack $\{\mathbf{t}^{(r)}_{ij}\}$ into matrix
    $\mathbf{T}^{(r)} \in \mathbb{R}^{B \times 2d}$

\Comment{Construct fake embeddings via noise injection}
\State Sample base Gaussian noise
    $\mathbf{N} \sim \mathcal{N}(0, 1)^{B \times 2d}$
\State Sample uniform perturbation
    $\boldsymbol{\epsilon} \sim \mathcal{U}(-1, 1)^{B \times 2d}$
\State Compute fake inputs
    $\mathbf{F} \gets \mathbf{N} + \tau \cdot \boldsymbol{\epsilon}$

\Comment{Generator forward pass}
\State $\hat{\mathbf{y}}^{(r)} \gets G(\mathbf{T}^{(r)})$
\State $\hat{\mathbf{y}}^{(f)} \gets G(\mathbf{F})$

\Comment{Discriminator forward pass}
\State $\hat{\mathbf{d}}^{(r)} \gets D(\hat{\mathbf{y}}^{(r)})$
\State $\hat{\mathbf{d}}^{(f)} \gets D(\hat{\mathbf{y}}^{(f)})$

\State \Return
    $\hat{\mathbf{y}}^{(r)}, \hat{\mathbf{y}}^{(f)},
     \hat{\mathbf{d}}^{(r)}, \hat{\mathbf{d}}^{(f)}$
\end{algorithmic}
\end{algorithm}

\textbf{Framework Analysis}: Interaction Learning via Adversarial Attention vs. CF

As discussed in the Introduction, real-world QoS observations are often corrupted by stochastic noise and outliers, which can mislead conventional latent factor and CF-based interaction models, as well as many graph-based approaches. Since these methods typically treat all observed interaction values as equally reliable, they are prone to fitting transient fluctuations and propagating such artifacts through their interaction functions, which in turn results in unstable and biased predictions under volatile service conditions.

The proposed AAIM module explicitly strengthens robustness at the interaction level by coupling a bidirectional hybrid attention generator with an adversarial discriminator. The dual-perspective attention design enables the generator to adaptively emphasize stable, consistently informative user–service patterns while suppressing features that exhibit irregular or noisy behavior, thus going beyond the largely static weighting schemes of traditional MF/CF-style models. Meanwhile, the discriminator is trained to distinguish realistic interaction scores from those obtained under Gumbel–Softmax–based perturbations, thereby forcing the generator to align its outputs with clean QoS signals in the discriminator’s view. This adversarial attention mechanism effectively filters out noisy and implausible interaction patterns, leading to more reliable and robust QoS prediction in noisy and rapidly changing Web service environments.

\subsection{Model Training}

After detailing the building blocks of QoSDiff, we now describe how the model is trained to perform full matrix completion. The training objective is formulated to jointly optimize the quality of QoS regression and the robustness of adversarial interaction learning.

Let $\mathcal{O} \subseteq \Omega$ denote the set of observed user--service pairs used for training.
For a mini-batch of interactions $(u_i, s_j) \in \mathcal{O}$ with normalized ground-truth QoS values $y_{ij}$, the generator (together with the upstream embedding and interaction modules) is optimized by a composite objective:
\begin{equation}
    \label{eq5.1.28}
    \mathcal{L}_G 
    = (1-\lambda\,) \mathcal{L}_{\mathrm{adv}}^{G}
      +  \lambda \, \mathcal{L}_{\mathrm{reg}},
\end{equation}
where $\lambda \in [0,1]$ controls the trade-off between adversarial supervision and QoS regression. 
The regression term is defined as
\begin{equation}
    \mathcal{L}_{\mathrm{reg}}
    = \frac{1}{|\mathcal{O}|}
      \sum_{(i,j)\in\mathcal{O}}
      \ell_{\mathrm{MSE}}\big(\hat{y}^{(r)}_{ij},\, y_{ij}\big),
\end{equation}
and the adversarial term for the generator is
\begin{equation}
    \mathcal{L}_{\mathrm{adv}}^{G}
    = \frac{1}{|\mathcal{O}|}
      \sum_{(i,j)\in\mathcal{O}}
      \ell_{\mathrm{BCE}}\big(\hat{d}^{(r)}_{ij},\, 1\big),
\end{equation}
where $\hat{y}^{(r)}_{ij}$ denotes the generator’s prediction for the real interaction embedding of $(u_i,s_j)$, and $\hat{d}^{(r)}_{ij}$ is the corresponding discriminator score. Here, the target label $1$ indicates that these scores should be classified as “real”.

The discriminator is trained to distinguish real from fake interaction scores via
\begin{equation}
    \label{eq5.1.29}
    \mathcal{L}_D 
    = \frac{1}{|\mathcal{O}|}
      \sum_{(i,j)\in\mathcal{O}}
      \Big[
        \ell_{\mathrm{BCE}}\big(\hat{d}^{(r)}_{ij},\, 1\big)
        + \ell_{\mathrm{BCE}}\big(\hat{d}^{(f)}_{ij},\, 0\big)
      \Big],
\end{equation}
where $\hat{d}^{(f)}_{ij}$ is the discriminator output for the fake interaction score generated from the Gumbel--Softmax–based embedding of $(u_i,s_j)$, and the target label $0$ marks these scores as “fake”.

For completeness, the point-wise losses are given by
\begin{equation}
    \label{eq5.1.30}
    \ell_{\mathrm{BCE}}(x, y)
    = - \Big[\, y \log\big(\sigma(x)\big)
              + (1 - y)\log\big(1 - \sigma(x)\big) \Big],
\end{equation}
where $\sigma(x) = 1 / \big(1 + \exp(-x)\big)$ is the logistic sigmoid function, and
\begin{equation}
    \label{eq5.1.31}
    \ell_{\mathrm{MSE}}(x, y) = (x - y)^2.
\end{equation}

In practice, the parameters of the discriminator and generator are updated in an alternating fashion with an update ratio of $1{:}1$: in each training iteration, we first minimize $\mathcal{L}_D$ with the generator fixed, and then minimize $\mathcal{L}_G$ with the discriminator frozen.

\section{Experinment}
\label{sec:exp}
In this section, we conduct extensive experiments to comprehensively evaluate the performance of the proposed framework. Specifically, we aim to answer the following research questions:

\begin{itemize}
    \item \textbf{RQ1 (Performance Superiority):} How does the proposed framework perform compared to state-of-the-art QoS prediction methods?
    \item \textbf{RQ2 (Cross-Dataset Generalization):} How well does the proposed model generalize to an additional QoS dataset with different characteristics?
    \item \textbf{RQ3 (Ablation Study):} How do different components contribute to the overall prediction performance?
    \item \textbf{RQ4 (Parameter Sensitivity):} How do key hyperparameters impact the model's effectiveness?
    \item \textbf{RQ5 (Robustness Analysis):} How robust is the proposed model against data noise and sparsity?
\end{itemize}

\subsection{Datasets and Evaluation Metrics}

\subsubsection{Datasets}
To evaluate the proposed framework in real-world scenarios, we conducted experiments on the widely used WS-DREAM dataset~\cite{WsDream}. Specifically, we utilized \textbf{Dataset 1}, which contains $1,974,675$ real-world QoS records describing the interactions between 339 users and 5,825 web services distributed globally. The dataset includes two key QoS properties: response time (RT) and throughput (TP).


\subsubsection{Evaluation Metrics}
We adopt two standard metrics, Mean Absolute Error (MAE) and Root Mean Square Error (RMSE), to measure the prediction accuracy. Let $\Omega_{\mathrm{test}} \subseteq \bar{\Omega}$ denote the set of user--service pairs in the test set, and let $|\Omega_{\mathrm{test}}|$ be its cardinality. For each $(i,j)\in\Omega_{\mathrm{test}}$, we denote by $y_{ij}$ and $\hat{y}_{ij}$ the ground-truth and predicted (normalized) QoS values, respectively. The metrics are defined as:
\begin{equation}
\mathrm{MAE} 
= \frac{1}{|\Omega_{\mathrm{test}}|}
  \sum_{(i,j)\in\Omega_{\mathrm{test}}}
  \left| y_{ij} - \hat{y}_{ij} \right|,
\end{equation}
\begin{equation}
\mathrm{RMSE} 
= \sqrt{ \frac{1}{|\Omega_{\mathrm{test}}|}
  \sum_{(i,j)\in\Omega_{\mathrm{test}}}
  \left( y_{ij} - \hat{y}_{ij} \right)^2 }.
\end{equation}
Lower MAE and RMSE values indicate better prediction accuracy.

\subsection{Experimental Setting}

\subsubsection{Implementation Details}
The proposed framework is implemented in PyTorch. 
All experiments are conducted on a workstation equipped with an Intel Core i7-12700H CPU @ 2.30\, GHz, 16\, GB RAM, and an NVIDIA GeForce RTX 3060 GPU, running Windows~11.

\subsubsection{Parameter Settings and Training Strategy}
We optimize both the diffusion-based generator and the adversarial discriminator using the AdamW optimizer\cite{loshchilov2017decoupled}. 
The maximum number of training epochs is set to 150, and we employ early stopping with a patience of 15 epochs based on the validation performance to avoid overfitting. 
Unless otherwise specified, the mini-batch size is fixed to 256 and the latent embedding dimension is set to 256, which provides a good trade-off between predictive accuracy and computational cost.%

Before training, we replace all missing entries in the raw QoS matrix (encoded as \(-1\)) with zeros and rescale all observed QoS values by dividing them by the global maximum. 
For each dataset with \(M\) users and \(N\) services and for each target density \(d \in \{2.5, 5, 7.5, 10\}\%\), 
we first collect all observed user--service pairs with non-zero QoS values and randomly permute their indices. 
We then sample \(\lfloor d \cdot M N \rfloor\) of these observed entries as the training set, 
a further \(\lfloor 0.05 \cdot M N \rfloor\) entries as the validation set, 
and use all remaining observed entries as the test set. 
This protocol ensures that the training, validation, and test sets are disjoint while the effective training density matches the desired sparsity level.

To mitigate the impact of random initialization and data shuffling, we repeat every experiment with three different random seeds and report the mean and standard deviation of all evaluation metrics over these runs.

\subsection{Baselines}

To comprehensively evaluate the performance of the proposed model, we select twelve representative baselines for comparison. These methods cover a broad spectrum of QoS prediction techniques, including CF-, MF-, deep-learning-, and GNN-based approaches. A brief description of each baseline is provided below:

\begin{itemize}
\item \textbf{UPCC} \cite{UPCC}: UPCC is a user-based collaborative filtering method that estimates missing QoS values by identifying the top-$k$ most similar users (measured via Pearson correlation) and aggregating their observed QoS scores.

\item \textbf{IPCC} \cite{IPCC}: IPCC is an item-based collaborative filtering method that predicts QoS values by exploiting the top-$k$ most similar services, thereby leveraging item-level relationships.

\item \textbf{UIPCC} \cite{UIPCC}: UIPCC is a hybrid CF method that integrates both user and service similarities, combining the principles of UPCC and IPCC to improve QoS prediction accuracy.

\item \textbf{PMF} \cite{PMF}: PMF is a probabilistic matrix factorization model that decomposes the observed QoS matrix into low-dimensional user and service latent factor matrices, which are then used to infer unobserved QoS values.

\item \textbf{BiasMF} \cite{BMF}: BiasMF extends standard matrix factorization by incorporating explicit bias terms to capture user- and service-specific tendencies, as well as global effects, thereby improving QoS prediction performance.

\item \textbf{CSMF} \cite{CSMF}: CSMF is a deep-learning-based model that performs context-aware embedding learning, extracting informative features from multiple contextual fields for QoS prediction.

\item \textbf{NFMF} \cite{NFMF}: NFMF is a deep-learning-based model that augments matrix factorization with fully connected networks and a multi-task learning framework, enabling it to jointly learn multiple QoS-related objectives.

\item \textbf{NCRL} \cite{NCRL}: NCRL is a deep-learning model with a dual-tower residual architecture. It employs multi-layer perceptrons to enhance feature representation and capture complex non-linear patterns in user–service interactions.

\item \textbf{GraphMF} \cite{GraphMF}: GraphMF is a GNN-based model that utilizes graph convolutional networks (GCNs) to explicitly encode user–service interaction graphs and capture high-order relational information for QoS prediction.

\item \textbf{PMP} \cite{he2024polarized}: PMP is a GNN-based framework that introduces a polarized message-passing paradigm, jointly exploiting similarities and differences between neighboring nodes to obtain dual information sources and enhance the expressive power of GNNs.

\item \textbf{RIGCN} \cite{wu2024robust}: RIGCN is a GNN-based architecture that incorporates outlier-aware pattern measurement into graph convolution to estimate user reputation. It further employs matrix factorization to model user–service interactions for QoS prediction.

\item \textbf{QoSGNN} \cite{QoSGNN}: QoSGNN is a GNN-based model that integrates attention mechanisms into graph neural architectures to dynamically prioritize informative user–service interactions and improve QoS prediction accuracy.
\end{itemize}

\subsection{Comparative Experimental Results (RQ1)}

\begin{table*}[]
\centering
\caption{Performance Comparison of QoS Prediction Models for Response Time on the WS-DREAM Dataset}

\label{tab:rt_comparison}
\resizebox{\textwidth}{!}{
\begin{tabular}{lcccccccc}
\hline
\multirow{2}{*}{\textbf{Model}} & \multicolumn{2}{c}{\textbf{Density=2.5\%}} & \multicolumn{2}{c}{\textbf{Density=5\%}} & \multicolumn{2}{c}{\textbf{Density=7.5\%}} & \multicolumn{2}{c}{\textbf{Density=10\%}} \\
\cline{2-9}
 & \textbf{MAE} & \textbf{RMSE} & \textbf{MAE} & \textbf{RMSE} & \textbf{MAE} & \textbf{RMSE} & \textbf{MAE} & \textbf{RMSE} \\
\hline
UPCC    & 0.709$\pm$0.007 & 1.467$\pm$0.008 & 0.640$\pm$0.021 & 1.380$\pm$0.003 & 0.588$\pm$0.003 & 1.339$\pm$0.003 & 0.556$\pm$0.003 & 1.309$\pm$0.004\\
IPCC    & 0.755$\pm$0.006 & 1.657$\pm$0.004 & 0.637$\pm$0.001 & 1.399$\pm$0.003 & 0.615$\pm$0.002 & 1.367$\pm$0.002 & 0.596$\pm$0.002 & 1.343$\pm$0.002\\
UIPCC   & 0.737$\pm$0.006 & 1.615$\pm$0.005 & 0.628$\pm$0.001 & 1.388$\pm$0.003 & 0.604$\pm$0.002 & 1.355$\pm$0.002 & 0.584$\pm$0.002 & 1.330$\pm$0.002\\
PMF     & 0.712$\pm$0.004 & 1.842$\pm$0.007 & 0.570$\pm$0.001 & 1.535$\pm$0.005 & 0.561$\pm$0.004 & 1.395$\pm$0.004 & 0.487$\pm$0.002 & 1.313$\pm$0.004\\
BiasMF  & 0.689$\pm$0.002 & 1.537$\pm$0.008 & 0.600$\pm$0.014 & 1.385$\pm$0.004 & 0.544$\pm$0.002 & 1.311$\pm$0.004 & 0.516$\pm$0.003 & 1.263$\pm$0.005\\
CSMF    & 0.649$\pm$0.003 & 1.678$\pm$0.009 & 0.550$\pm$0.001 & 1.494$\pm$0.002 & 0.497$\pm$0.002 & 1.407$\pm$0.003 & 0.453$\pm$0.001 & 1.356$\pm$0.001\\
NFMF    & 0.524$\pm$0.004 & 1.484$\pm$0.031 & 0.447$\pm$0.005 & 1.355$\pm$0.010 & 0.426$\pm$0.001 & 1.322$\pm$0.006 & 0.413$\pm$0.003 & 1.303$\pm$0.004\\
NCRL    & 0.561$\pm$0.005 & 1.591$\pm$0.019 & 0.546$\pm$0.002 & 1.564$\pm$0.016 & 0.542$\pm$0.004 & 1.540$\pm$0.015 & 0.537$\pm$0.004 & 1.519$\pm$0.007\\
GraphMF & 0.448$\pm$0.005 & 1.415$\pm$0.011 & 0.399$\pm$0.007 & 1.345$\pm$0.012 & 0.378$\pm$0.005 & 1.302$\pm$0.009 & 0.367$\pm$0.003 & 1.299$\pm$0.008\\
PMP     & 0.538$\pm$0.003 & 1.510$\pm$0.003 & 0.465$\pm$0.002 & 1.438$\pm$0.003 & 0.436$\pm$0.004 & 1.412$\pm$0.003 & 0.418$\pm$0.002 & 1.387$\pm$0.004\\
RIGCN   & 0.497$\pm$0.001 & 1.511$\pm$0.008 & 0.435$\pm$0.009 & 1.410$\pm$0.033 & 0.419$\pm$0.012 & 1.457$\pm$0.011 & 0.395$\pm$0.001 & 1.452$\pm$0.008\\
QoSGNN  & 0.431$\pm$0.005 & 1.439$\pm$0.009 & 0.377$\pm$0.006 & 1.335$\pm$0.008 & 0.353$\pm$0.004 & 1.295$\pm$0.011 & 0.345$\pm$0.004 & 1.276$\pm$0.010\\ \hline
QoSDiff & 0.402$\pm$0.004 & 1.394$\pm$0.006 & 0.358$\pm$0.006 & 1.330$\pm$0.002 & 0.337$\pm$0.001 & 1.284$\pm$0.006 & 0.324$\pm$0.001 & 1.253$\pm$0.005\\
\textit{imp.} 
& \textit{6.73\%} & \textit{3.13\%}
& \textit{5.04\%} & \textit{0.37\%}
& \textit{4.53\%} & \textit{0.85\%}
& \textit{6.09\%} & \textit{1.80\%}\\
\hline
\end{tabular}
}
\end{table*}

\begin{table*}[]
\centering
\caption{Performance Comparison of QoS Prediction Models for Throughput on the WS-DREAM Dataset}
\label{tab:tp_comparison}
\resizebox{\textwidth}{!}{
\begin{tabular}{lcccccccc}
\hline
\multirow{2}{*}{\textbf{Model}} & \multicolumn{2}{c}{\textbf{Density=2.5\%}} & \multicolumn{2}{c}{\textbf{Density=5\%}} & \multicolumn{2}{c}{\textbf{Density=7.5\%}} & \multicolumn{2}{c}{\textbf{Density=10\%}} \\
\cline{2-9}
 & \textbf{MAE} & \textbf{RMSE} & \textbf{MAE} & \textbf{RMSE} & \textbf{MAE} & \textbf{RMSE} & \textbf{MAE} & \textbf{RMSE} \\
\hline
UPCC    & $31.759\pm0.166$ & $67.816\pm0.221$ & $27.209\pm0.159$ & $60.907\pm0.179$ & $24.492\pm0.126$ & $57.176\pm0.039$ & $22.605\pm0.085$ & $54.522\pm0.113$ \\
IPCC    & $31.781\pm0.283$ & $73.172\pm0.515$ & $27.769\pm0.129$ & $62.929\pm0.126$ & $26.415\pm0.079$ & $61.318\pm0.067$ & $26.182\pm0.053$ & $60.353\pm0.061$ \\
UIPCC   & $31.260\pm0.160$ & $67.397\pm0.221$ & $26.749\pm0.143$ & $60.683\pm0.175$ & $24.166\pm0.118$ & $57.041\pm0.043$ & $22.364\pm0.077$ & $54.421\pm0.112$ \\
PMF     & $24.287\pm0.280$ & $72.125\pm0.844$ & $19.078\pm0.147$ & $57.814\pm0.508$ & $17.024\pm0.135$ & $51.516\pm0.423$ & $16.027\pm0.073$ & $48.256\pm0.246$ \\
BiasMF  & $28.618\pm0.320$ & $69.494\pm0.861$ & $21.816\pm0.195$ & $56.766\pm0.510$ & $19.260\pm0.108$ & $51.277\pm0.296$ & $17.883\pm0.139$ & $48.390\pm0.263$ \\
CSMF    & $26.157\pm0.254$ & $72.285\pm0.584$ & $20.824\pm0.085$ & $58.806\pm0.586$ & $18.236\pm0.082$ & $52.563\pm0.374$ & $16.691\pm0.062$ & $48.996\pm0.354$ \\
NFMF    & $25.878\pm0.289$ & $67.975\pm1.737$ & $20.086\pm0.307$ & $54.626\pm0.487$ & $18.252\pm0.368$ & $51.802\pm0.957$ & $17.602\pm0.222$ & $50.900\pm0.970$ \\
NCRL    & $28.418\pm0.443$ & $80.555\pm1.702$ & $27.130\pm0.414$ & $77.011\pm0.785$ & $26.867\pm0.298$ & $75.953\pm1.127$ & $26.648\pm0.230$ & $75.413\pm0.502$ \\
GraphMF & $19.514\pm0.745$ & $59.849\pm1.093$ & $16.140\pm1.100$ & $52.489\pm2.621$ & $15.648\pm0.901$ & $50.953\pm2.182$ & $15.287\pm0.833$ & $50.171\pm2.322$ \\
PMP     & $23.884\pm0.185$ & $81.198\pm1.216$ & $18.240\pm0.080$ & $71.558\pm0.444$ & $16.396\pm0.064$ & $67.900\pm0.191$ & $15.425\pm0.075$ & $66.160\pm0.404$ \\
RIGCN   & $21.854\pm1.189$ & $72.790\pm1.132$ & $18.741\pm1.137$ & $65.978\pm2.606$ & $17.325\pm1.042$ & $63.953\pm2.338$ & $16.164\pm1.028$ & $65.495\pm2.255$ \\
QoSGNN  & $18.823\pm0.045$ & $59.303\pm0.190$ & $16.255\pm0.116$ & $54.206\pm0.188$ & $14.489\pm0.359$ & $48.640\pm1.013$ & $13.946\pm0.355$ & $47.955\pm0.756$ \\ \hline
QoSDiff & $16.621\pm0.200$ & $55.588\pm0.982$ & $13.473\pm0.061$ & $47.283\pm0.528$ & $12.170\pm0.041$ & $43.8096\pm0.222$ & $11.474\pm0.069$ & $41.781\pm0.231$ \\ 

\textit{imp.} 
& \textit{11.70\%} & \textit{6.26\%}
& \textit{17.11\%} & \textit{12.77\%}
& \textit{16.01\%} & \textit{9.93\%}
& \textit{17.73\%} & \textit{12.87\%}\\\hline

\end{tabular}
}
\end{table*}

Tables \ref{tab:rt_comparison} and \ref{tab:tp_comparison} illustrate the predictive performance of various methods on non-functional QoS attributes (response time and throughput) under four different matrix densities (2.5\%, 5\%, 7.5\%, 10\%). Experimental results obtained with three different random seeds are presented in terms of the mean and standard deviation of MAE and RMSE. From these results, we draw the following conclusions:

\begin{itemize}
   \item CF-based methods such as UPCC, IPCC, and UIPCC primarily predict missing QoS values through direct similarity calculations among users and services. The results indicate these methods perform poorly at low data densities, demonstrating significant sensitivity to data sparsity, particularly in the 2.5\% density condition.
    \item  MF-based methods (PMF and BiasMF) effectively capture latent features of users and services, especially under higher data densities. However, these methods are highly sensitive to the selection of matrix density, resulting in notable performance fluctuations under sparse or complex data conditions. For example, the unstable performance of PMF, particularly in RMSE metrics, highlights this drawback.
    \item DL-based methods (CSMF, NFMF, and NCRL) extend the modeling capabilities by exploring nonlinear and high-order interactions. The experimental results indicate that NFMF generally outperforms other DL methods due to its multi-task learning strategy, allowing for the effective extraction of latent user-service features. Conversely, NCRL underperforms consistently and shows low sensitivity to matrix density variations, indicating that embedding learning requires a well-structured framework.
    \item GNN-based methods (GraphMF, PMP, RIGCN, QoSGNN).
    By explicitly exploiting user–service graph structures, these models generally achieve better performance than purely neural baselines. 
    However, GraphMF relies on a shallow GCN backbone whose limited expressiveness constrains the quality of the learned embeddings. 
    PMP introduces a novel, general-purpose message-passing strategy, yet our results on both datasets suggest that such a generic scheme is not particularly effective for domain-specific representation learning in service computing. 
    RIGCN leverages OPM to model user reputation, but its dependence on hand-crafted statistical features restricts its predictive capacity. 
 QoSGNN, the strongest baseline, incorporates attention mechanisms but is still bound by the quality of the constructed graph. Under extreme sparsity (2.5\%), the explicit edges are insufficient for effective message passing, leading to performance saturation.

    \item  Our proposed framework achieves the best performance across all density settings and metrics. Notably, the improvement is most substantial at the lowest density (2.5\%), as indicated by the 6.73\% gain in MAE for Response Time. This validates that QoSDiff is less dependent on explicit, noisy edge connections. By replacing rigid message passing with a global diffusion-based embedding generation, our model effectively infer plausible latent features even with minimal observations, demonstrating superior robustness and generalization capability in sparse service environments.
    
\end{itemize}

\subsection{Cross-Dataset Generalization (RQ2)}
To assess the cross-dataset generalization ability of QoSDiff, we further conduct experiments on the EEL dataset~\cite{10660479}, a large-scale edge–cloud latency corpus built from nearly 900 million PING measurements collected over 5174 edge nodes, representing one of the largest edge deployments reported so far. 
Unlike WS-DREAM, where QoS is defined between end users and Web services, EEL specifies two non-functional attributes for each pair of edge nodes: the end-to-end latency \texttt{DELAY} and the hop count \texttt{HOPS} between the two nodes. 
In our setting, one node is treated as the user and the other as the service, and we construct corresponding user–service QoS matrices for both latency and hop count.

Given the substantial scale of EEL, we select a subset of strong and representative baselines for comparison and construct user–service latency matrices under three sparsity levels (2.5\%, 5\%, and 7.5\%). 
To accelerate convergence on this large corpus, we slightly adjust the mini-batch size and set it to $8192$ for all methods on EEL.
Apart from this change, the training protocol, evaluation metrics, and remaining hyperparameter configurations are kept identical to those used in the main WS-DREAM experiments to ensure a fair cross-dataset comparison. 
The results for \texttt{DELAY} and \texttt{HOPS} are summarized in Tables~\ref{tab:delay_comparison_new} and~\ref{tab:hops_comparison_newer}, respectively.

\begin{table*}[]
\centering
\caption{Performance Comparison of QoS Prediction Models for DELAY on the EEL Dataset}

\label{tab:delay_comparison_new}
\resizebox{\textwidth}{!}{%
\begin{tabular}{lcccccc}
\hline
\multirow{2}{*}{\textbf{Model}} & \multicolumn{2}{c}{\textbf{Density=2.5\%}} & \multicolumn{2}{c}{\textbf{Density=5\%}} & \multicolumn{2}{c}{\textbf{Density=7.5\%}} \\
\cline{2-7}
& \textbf{MAE} & \textbf{RMSE} & \textbf{MAE} & \textbf{RMSE} & \textbf{MAE} & \textbf{RMSE} \\
\hline
PMF      & 0.008741$\pm$0.001708 & 0.020204$\pm$0.001106 & 0.007073$\pm$0.000041 & 0.019539$\pm$0.000051 & 0.007009$\pm$0.000084 & 0.018848$\pm$0.000058 \\
CSMF     & 0.008085$\pm$0.000020 & 0.020096$\pm$0.000333 & 0.007315$\pm$0.000020 & 0.018446$\pm$0.000136 & 0.006930$\pm$0.000029 & 0.017475$\pm$0.000053 \\
NFMF     & 0.007702$\pm$0.000052 & 0.019906$\pm$0.000152 & 0.007132$\pm$0.000098 & 0.018497$\pm$0.000764 & 0.006789$\pm$0.000095 & 0.017214$\pm$0.000054 \\
NCRL     & 0.007770$\pm$0.000179 & 0.019038$\pm$0.000256 & 0.007236$\pm$0.000166 & 0.018585$\pm$0.000617 & 0.006869$\pm$0.000087 & 0.017736$\pm$0.000175 \\
GraphMF  & 0.007245$\pm$0.000102 & 0.020375$\pm$0.000586 & 0.007012$\pm$0.000135 & 0.018969$\pm$0.000239 & 0.006497$\pm$0.000192 & 0.017971$\pm$0.000383 \\
PMP      & 0.007369$\pm$0.000102 & 0.018181$\pm$0.000648 & 0.006918$\pm$0.000062 & 0.018156$\pm$0.000108 & 0.006540$\pm$0.000029 & 0.017872$\pm$0.000135 \\
QoSGNN   & 0.007199$\pm$0.000582 & 0.019158$\pm$0.000794 & 0.006588$\pm$0.000322 & 0.018664$\pm$0.000239 & 0.005882$\pm$0.000246 & 0.017478$\pm$0.000345 \\ 

\hline
QoSDiff  & 0.006842$\pm$0.000253 & 0.018645$\pm$0.000896 & 0.006177$\pm$0.000150 & 0.018321$\pm$0.000581 & 0.005581$\pm$0.000090 & 0.016322$\pm$0.000123 \\
\textit{imp.} & \textit{4.96\%} & \textit{-2.55\%} & \textit{6.24\%} & \textit{-0.91\%} & \textit{5.12\%} & \textit{5.18\%} \\

\hline
\end{tabular}
}
\end{table*}

\begin{table*}[]
\centering
\caption{Performance Comparison of QoS Prediction Models for HOPS on the EEL Dataset}
\label{tab:hops_comparison_newer}
\resizebox{\textwidth}{!}{%
\begin{tabular}{lcccccc}
\hline
\multirow{2}{*}{\textbf{Model}} & \multicolumn{2}{c}{\textbf{Density=2.5\%}} & \multicolumn{2}{c}{\textbf{Density=5\%}} & \multicolumn{2}{c}{\textbf{Density=7.5\%}} \\
\cline{2-7}
& \textbf{MAE} & \textbf{RMSE} & \textbf{MAE} & \textbf{RMSE} & \textbf{MAE} & \textbf{RMSE} \\
\hline
PMF        & 0.021827$\pm$0.001262 & 0.027745$\pm$0.001305 & 0.018163$\pm$0.002771 & 0.024393$\pm$0.002592 & 0.015867$\pm$0.003417 & 0.021887$\pm$0.003155 \\
CSMF       & 0.008832$\pm$0.000083 & 0.013039$\pm$0.000068 & 0.007227$\pm$0.000545 & 0.012059$\pm$0.000401 & 0.006009$\pm$0.000042 & 0.010188$\pm$0.000020 \\
NFMF       & 0.007270$\pm$0.000041 & 0.012088$\pm$0.000258 & 0.006070$\pm$0.000059 & 0.011990$\pm$0.000124 & 0.005242$\pm$0.000049 & 0.010092$\pm$0.000183 \\
NCRL       & 0.007964$\pm$0.000123 & 0.012513$\pm$0.000100 & 0.006103$\pm$0.000136 & 0.011590$\pm$0.000076 & 0.005344$\pm$0.000093 & 0.009979$\pm$0.000045 \\
GraphMF    & 0.006679$\pm$0.000302 & 0.012050$\pm$0.000337 & 0.005644$\pm$0.000106 & 0.011812$\pm$0.000063 & 0.004921$\pm$0.0000982 & 0.010009$\pm$0.000150\\
PMP        & 0.008842$\pm$0.000932 & 0.013775$\pm$0.000941 & 0.006359$\pm$0.000358 & 0.011954$\pm$0.000338 & 0.004917$\pm$0.000104 & 0.010185$\pm$0.000307\\
QoSGNN     & 0.006663$\pm$0.000482 & 0.011607$\pm$0.000222 & 0.005330$\pm$0.000161 & 0.011480$\pm$0.000057 & 0.004831$\pm$0.000235 & 0.010091$\pm$0.000064\\ \hline
QoSDiff & 0.00529$\pm$0.000341 & 0.010210$\pm$0.000105 & 0.004848$\pm$0.000117 & 0.010804$\pm$0.000132 & 0.003925$\pm$0.000212 & 0.009384$\pm$0.000111 \\
\textit{imp.} & \textit{20.61\%} & \textit{12.04\%} & \textit{9.04\%} & \textit{5.89\%} & \textit{18.75\%} & \textit{5.96\%} \\

\hline
\end{tabular}
}
\end{table*}

From Tables~\ref{tab:delay_comparison_new} and~\ref{tab:hops_comparison_newer}, several observations can be made. 
First, QoSDiff consistently achieves the lowest MAE across all sparsity levels for both \texttt{DELAY} and \texttt{HOPS}, and yields clear RMSE improvements in most cases, indicating that the proposed diffusion-guided embedding learning and adversarial attention module transfer well from WS-DREAM to the large-scale EEL environment. 
This demonstrates that QoSDiff exhibits strong cross-dataset generalization ability, even when the underlying QoS semantics and interaction patterns differ from those in the original training corpus.

Second, on the \texttt{DELAY} prediction task, PMP attains slightly better RMSE than all methods, including QoSDiff, at 2.5\% and 5\% densities, suggesting that carefully designed message-passing mechanisms can still be highly effective when the underlying graph structure is reliable and the prediction target is purely latency-oriented. 
However, PMP does not exhibit the same advantage on the \texttt{HOPS} prediction task and tends to be less stable as sparsity decreases, whereas QoSDiff maintains robust gains across both metrics. 
These results imply that although PMP can deliver strong performance under specific conditions, its applicability to broader service-computing scenarios is more limited, while QoSDiff offers a more generally robust solution across heterogeneous QoS datasets.

\subsection{Ablation Study (RQ3)}

To further examine the contribution of each core component in the proposed framework, we design two groups of ablation experiments focusing on (i) embedding learning strategies and (ii) interaction modeling mechanisms. 
All experiments are conducted on the RT and TP datasets under four matrix densities(2.5\%, 5\%, 7.5\%, and 10\%).

\subsubsection{Embedding Learning Methods}

To verify the effectiveness of the diffusion-based embedding learning module in capturing high-order relationships between users and services, we compare a vanilla embedding variant with the full diffusion-guided variant. 
In the vanilla setting, user and service representations are obtained by directly aggregating identity and context embeddings without any diffusion refinement. 
In the full setting, these initial embeddings are further processed by the proposed denoising diffusion module.

\begin{figure*}[]
    \centering
    \subfigure{
        \includegraphics[width=0.48\linewidth]{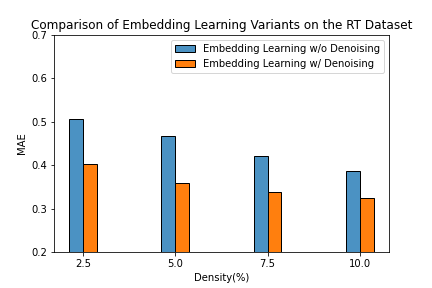}
    }
    \subfigure{
        \includegraphics[width=0.48\linewidth]{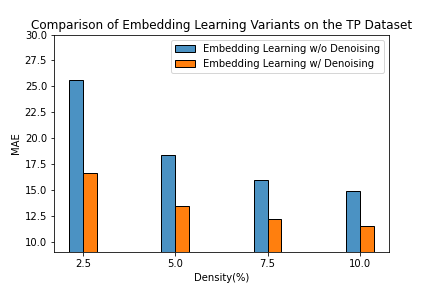}
    }

    \caption{Model performance comparison of different embedding learning methods}
    \label{Embedding}
\end{figure*}

Figure~\ref{Embedding} reports the results on the RT and TP datasets under different matrix densities. 
Across all sparsity levels, the variant equipped with denoising consistently achieves lower MAE than the vanilla embedding version, demonstrating that the diffusion-guided refinement effectively enhances the quality of latent representations. 
This suggests that modeling the denoising process over the embedding space helps the framework capture more informative and stable user--service relations, which in turn leads to more accurate QoS predictions.

\subsubsection{Interaction Learning Methods}

The interaction module is responsible for modeling latent dependencies between users and services based on their learned embeddings. 
To assess the effectiveness of the proposed adversarial attention-based interaction module (AAIM), we consider three interaction strategies within the same framework: (i) simple inner product, corresponding to a matrix factorization (MF) style interaction; (ii) a multi-layer perceptron, representing neural collaborative filtering (CF); and (iii) the proposed AAIM.

\begin{figure*}[]
    \centering
    \subfigure{
        \includegraphics[width=0.48\linewidth]{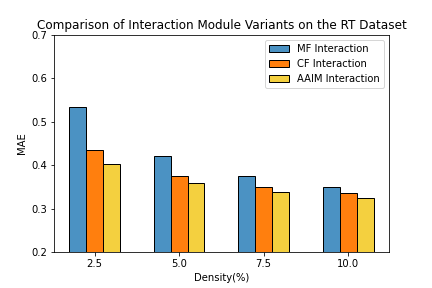}
    }
    \subfigure{
        \includegraphics[width=0.48\linewidth]{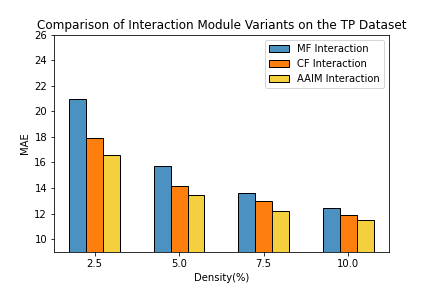}
    }
    \caption{Model performance comparison of different interaction learning methods}
    \label{Int}
\end{figure*}

As shown in Figure~\ref{Int}, the AAIM-based variant consistently attains the lowest MAE across all density levels and on the WS-Dream datasets. 
Compared with inner-product and standard neural interaction functions, the self-attention-based hybrid interaction module can better capture complex, high-order user--service relations and suppress noisy patterns through adversarial training, thereby providing more accurate QoS predictions.

\subsection{Hyperparameter Analysis (RQ4)}

To explore the influence of hyperparameters on model performance, we conducted comprehensive experiments focusing on embedding dimensions and the number of attention heads in the denoising diffusion stage. MAE and RMSE are used as the evaluation metric across both RT and TP datasets at matrix densities of 2.5\%, 5\%, 7.5\%, and 10\%.

\subsubsection{Embedding Dimension}

\begin{figure*}[htbp]
\centering
\includegraphics[width = \textwidth]{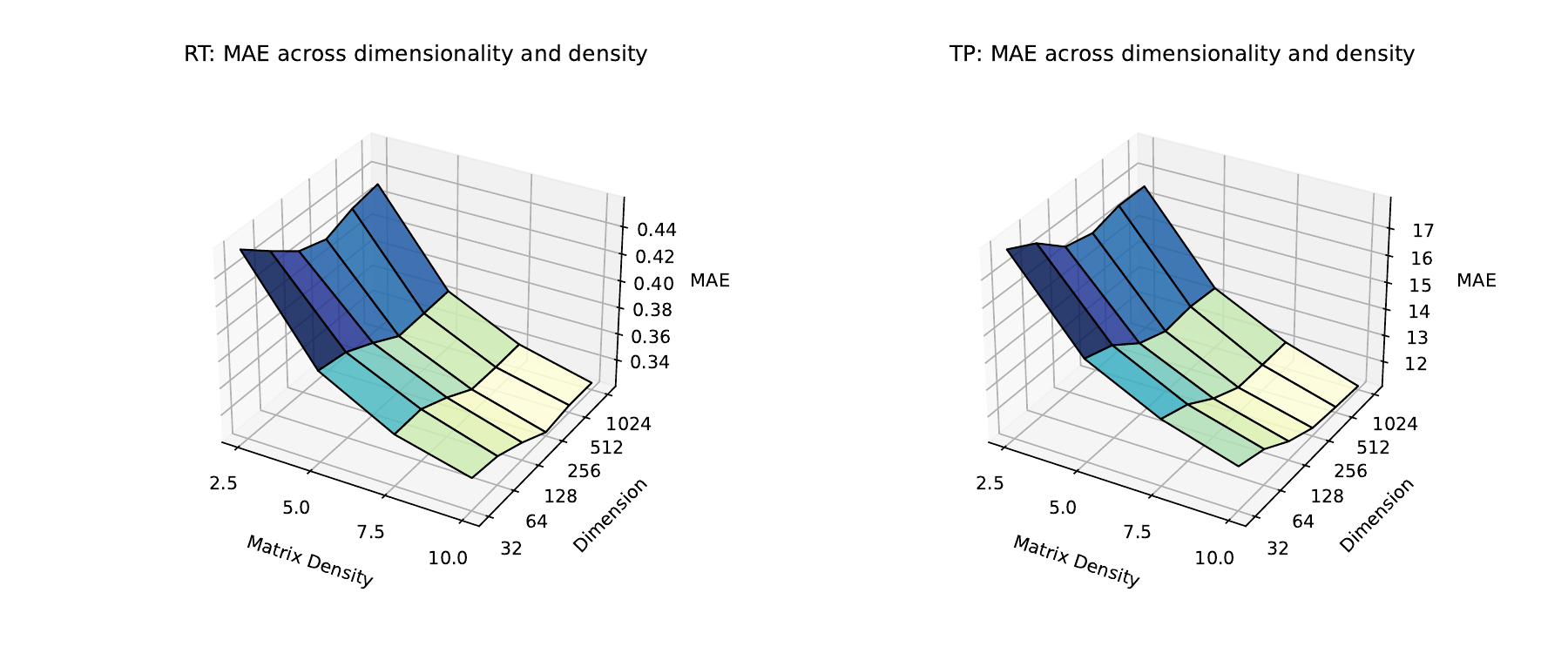}
\caption{Model performance across different embedding dimensions}
\label{dimension}
\end{figure*}


In our framework, the embedding dimension maps nodes to a higher-dimensional space, enabling the model to capture more intricate data patterns. The experimental results are presented in Figure \ref{dimension}:

The results indicate an initial decrease followed by a subsequent increase in prediction performance as the embedding dimensions increase. This pattern likely arises because sufficient dimensionality is required for the denoising diffusion process to effectively represent matrices approximating Gaussian distributions. Embedding dimensions of 256 or higher fulfill this condition and yield optimal performance. Consequently, we selected an embedding dimension of 256 for subsequent experiments.

\subsubsection{Number of Attention Heads}


\begin{figure*}[htbp]
\centering
\includegraphics[width = \textwidth]{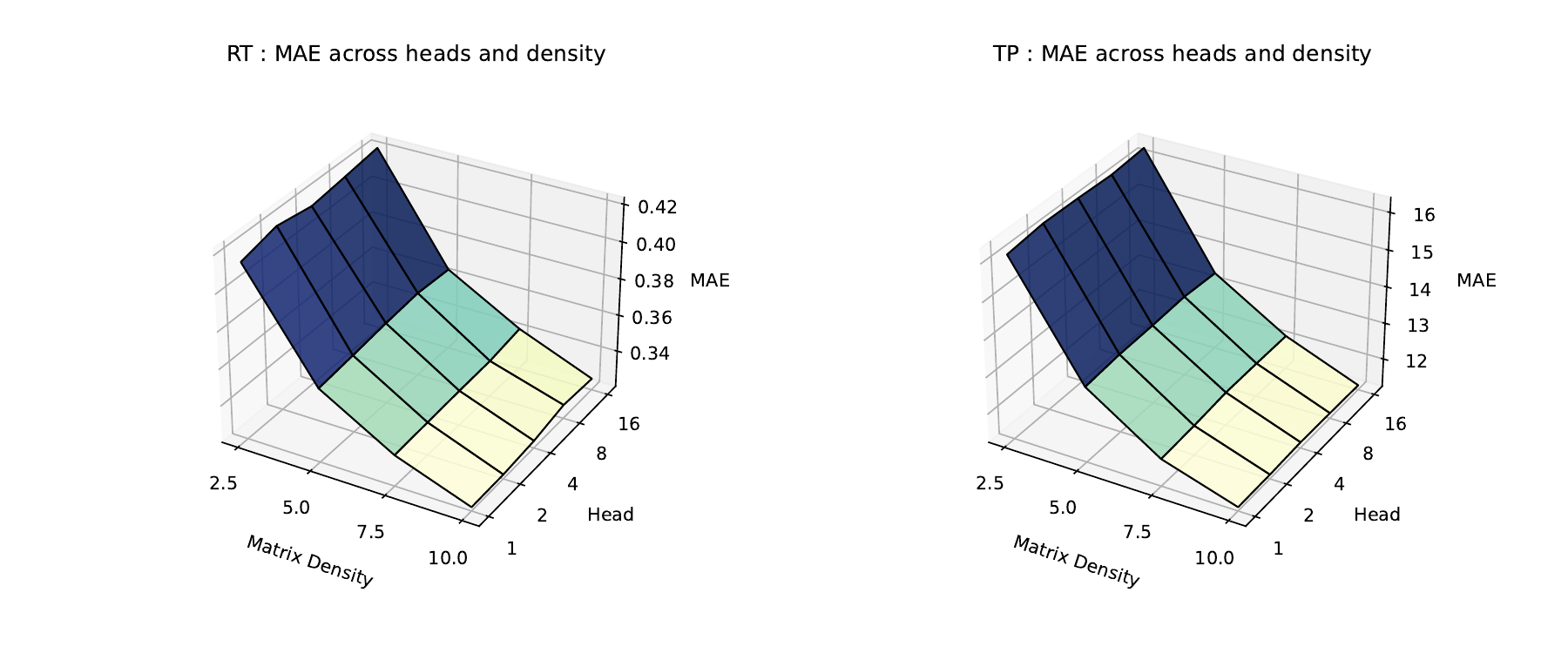}
\caption{Model performance across different attention heads}
\label{head_num}
\end{figure*}

The multi-head attention mechanism enhances the noise prediction capability of the transformed matrices by concurrently modeling diverse noise patterns throughout the diffusion process. To assess its impact, we vary the number of attention heads as {1, 2, 4, 8, 16}. The corresponding results are reported in Figure~\ref{head_num}.

Experimental findings reveal a deterioration in model performance as the number of attention heads increases. This suggests that a single-head attention mechanism effectively focuses on critical latent relationships within matrices, while additional heads may lead to both the dispersion of attention focus and the increased model complexity, negatively affecting feature extraction. Thus, we set the number of attention heads to one in the main experiments to ensure optimal predictive performance.

\subsubsection{Trade-off Parameter $\lambda$ between adversarial and regression losses}

\begin{figure*}[htbp]
\centering
\includegraphics[width = \textwidth]{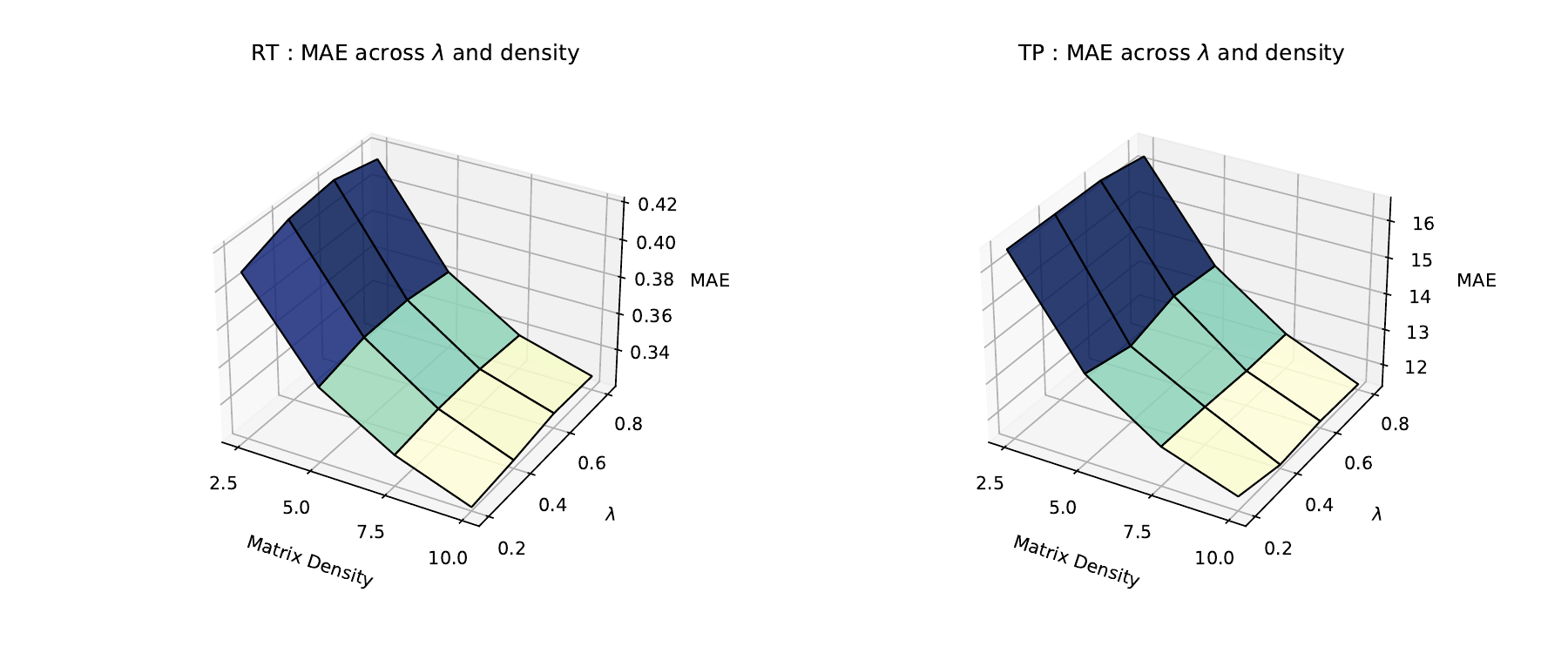}
\caption{Model performance across different $\lambda$}
\label{lmd}
\end{figure*}

The coefficient $\lambda$ in the overall training objective (cf. Eq.~(\ref{eq5.1.28})) acts as a trade-off controller between the adversarial loss and the regression loss. A proper choice of $\lambda$ is therefore crucial for balancing the stability of adversarial training and the accuracy of QoS regression. To investigate its influence, we conduct a sensitivity study by varying $\lambda \in \{0.2, 0.4, 0.6, 0.8\}$ on both the RT and TP datasets under four matrix densities (2.5\%, 5\%, 7.5\%, and 10\%). The corresponding results are summarized in Fig.~\ref{lmd}.

From Fig.~\ref{lmd}, we observe that the optimal setting of $\lambda$ exhibits different trends on the two datasets. On the RT dataset, setting $\lambda=0.2$ consistently achieves the lowest MAE across all density levels, indicating that the regression term should dominate the optimization and that an overly strong adversarial signal may be unnecessary. In contrast, on the TP dataset, the best performance at 2.5\% density is obtained when $\lambda=0.8$, whereas $\lambda=0.4$ performs best for the other three densities. This suggests that TP benefits more from a relatively stronger adversarial regularization.

Considering the trade-off between performance and simplicity, we set $\lambda=0.2$ as a unified configuration. 
This choice is clearly optimal on RT and only slightly worse (within 2.6\%) than the best settings on TP.

\subsection{Robustness Test of Our Proposed Model Against Data Noise (RQ5)} 

In this section, we investigate the robustness of QoSDiff under different levels of observational noise. The dataset is first split into training, validation, and test sets. To evaluate robustness against corrupted observations, we keep the training and validation data clean and inject synthetic noise only into the test interactions.

Let $\Omega_{\mathrm{test}} \subseteq \bar{\Omega}$ denote the index set of user--service pairs in the test set, and let $N_{\mathrm{test}} = |\Omega_{\mathrm{test}}|$. Each test interaction is represented as a triplet $(u_i, s_j, y_{ij})$ with $(i,j)\in\Omega_{\mathrm{test}}$, where $y_{ij}$ is the normalized QoS value. To simulate mislogged or mismatched QoS feedback, we randomly select a subset $\Omega_p \subseteq \Omega_{\mathrm{test}}$ containing $p\%$ of the test entries. 

For each $(i,j) \in \Omega_p$, we randomly perturb the user and service identities while keeping the QoS value unchanged. Concretely, we sample
\begin{equation}
    u' \sim \mathrm{Unif}\big(\{1,\dots,m\}\big), \qquad
    s' \sim \mathrm{Unif}\big(\{1,\dots,n\}\big),
\end{equation}
and replace the original triplet $(u_i, s_j, y_{ij})$ with $(u', s', y_{ij})$ in the corrupted test set. This process simulates deployment-time logging errors where observed QoS values are attached to incorrect user--service pairs.

We compare QoSDiff with QoSGNN under this noisy evaluation protocol. Both models adopt embedding-based representations to capture latent user and service characteristics. QoSGNN refines these embeddings via GNN-based message passing and models user--service interactions with conventional multi-layer perceptrons. In contrast, QoSDiff employs an inverse diffusion probabilistic embedding module to learn robust latent representations without relying on explicit graph construction, and further leverages a bidirectional user--service interaction attention mechanism to capture high-order dependencies between users and services.
We thus focus our robustness comparison on QoSGNN as the strongest GNN-based baseline.

Experiments are conducted on the RT and TP datasets under four matrix densities (2.5\%, 5\%, 7.5\%, and 10\%). For each setting, we report MAE and RMSE for both QoSDiff and QoSGNN, and the column $Degrad.$ denotes the relative error increase compared with the noise-free case (0\% noise), 
computed as $(\mathrm{MAE}_{p} - \mathrm{MAE}_{0}) / \mathrm{MAE}_{0} \times 100\%$ for a noise ratio $p$.
 The detailed results are summarized in Tables~\ref{tab:robustness-mae-rt}--\ref{tab:robustness-rmse-tp}.

\begin{table*}[htbp]
\centering
\caption{MAE Comparison and Growth Rate under Different Noise Levels and Densities on RT Datasets}
\resizebox{\textwidth}{!}{
\begin{tabular}{c|cc|cc|cc|cc|cc|cc|cc|cc}
\toprule
\multirow{2}{*}{Noise (\%)} 
& \multicolumn{2}{c|}{QoSGNN@2.5} & \multicolumn{2}{c|}{QoSDiff@2.5} 
& \multicolumn{2}{c|}{QoSGNN@5} & \multicolumn{2}{c|}{QoSDiff@5}
& \multicolumn{2}{c|}{QoSGNN@7.5} & \multicolumn{2}{c|}{QoSDiff@7.5}
& \multicolumn{2}{c|}{QoSGNN@10} & \multicolumn{2}{c|}{QoSDiff@10} \\
& MAE & Degrad. & MAE & Degrad. & MAE & Degrad. & MAE & Degrad. 
& MAE & Degrad. & MAE & Degrad. & MAE & Degrad. & MAE & Degrad. \\
\midrule
0  & 0.4459 & --   & 0.4139 & --   & 0.3826 & --           & 0.3610 & --   &  0.3490 & --   & 0.3390 & --   & 0.3455 & --   & 0.3242 & --   \\
5  & 0.4892 & 9.7\%  & 0.4493 & 8.55\%  & 0.4351 & 13.7\% & 0.4006 & 10.97\% & 0.3959 & 13.4\% & 0.3795 & 11.95\% & 0.3851 & 11.5\% & 0.3560 & 9.81\% \\
10 & 0.5320 & 19.3\% & 0.4826 & 16.6\% & 0.4859 & 27.0\% & 0.4534 & 25.6\% & 0.4532 & 29.9\% & 0.4303 & 26.93\% & 0.4419 & 27.9\% & 0.3985 & 22.92\% \\
15 & 0.5736 & 28.6\% & 0.5217 & 26.04\% & 0.5376 & 40.5\% & 0.5019 & 39.03\% & 0.5063 & 45.1\% & 0.4812 & 41.95\% & 0.4984 & 44.3\% &  0.4641 & 43.15\% \\
20 & 0.6151 & 37.9\% & 0.5537& 33.78\% & 0.5875 & 53.6\% & 0.5448 & 50.91\% & 0.5613 & 60.8\% & 0.5276 & 55.63\% & 0.5564 & 61.0\% & 0.4910 & 51.45\% \\
25 & 0.6558 & 47.0\% & 0.6018 & 45.4\% & 0.6347 & 65.9\% &0.5711  & 58.2\% & 0.6123 & 75.5\% & 0.5801 & 71.12\% & 0.6058 & 75.3\% & 0.5294 & 63.29\% \\
\bottomrule
\end{tabular}}
\label{tab:robustness-mae-rt}
\end{table*}

\begin{table*}[htbp]
\centering
\caption{RMSE Comparison and Growth Rate under Different Noise Levels and Densities on RT Datasets}
\resizebox{\textwidth}{!}{
\begin{tabular}{c|cc|cc|cc|cc|cc|cc|cc|cc}
\toprule
\multirow{2}{*}{Noise (\%)} 
& \multicolumn{2}{c|}{QoSGNN@2.5} & \multicolumn{2}{c|}{QoSDiff@2.5} 
& \multicolumn{2}{c|}{QoSGNN@5} & \multicolumn{2}{c|}{QoSDiff@5}
& \multicolumn{2}{c|}{QoSGNN@7.5} & \multicolumn{2}{c|}{QoSDiff@7.5}
& \multicolumn{2}{c|}{QoSGNN@10} & \multicolumn{2}{c|}{QoSDiff@10} \\
& RMSE & Degrad. & RMSE & Degrad. & RMSE & Degrad. & RMSE & Degrad. 
& RMSE & Degrad. & RMSE & Degrad. & RMSE & Degrad. & RMSE & Degrad. \\
\midrule
0  & 1.4290 & --      & 1.4070 & --      & 1.3341 & --      & 1.3285 & --      & 1.2918 & --      & 1.2785 & --      & 1.2718 & --      & 1.2530 & --   \\
5  & 1.4985 & 7.00\%  & 1.5909 & 5.62\%  & 1.4533 & 8.93\%  & 1.4352 & 8.03\%  & 1.4045 & 8.72\%  & 1.3668 & 6.91\%  & 1.3815 & 8.91\%  & 1.3437 & 7.24\% \\
10 & 1.5899 & 13.52\% & 1.6358 & 13.07\% & 1.5573 & 16.73\% & 1.5190 & 14.34\% & 1.5298 & 18.42\% & 1.4811 & 15.85\% & 1.509  & 18.65\% & 1.4441 & 15.25\% \\
15 & 1.6734 & 19.49\% & 1.7191 & 16.26\% & 1.6347 & 24.64\% & 1.6202 & 21.96\% & 1.6347 & 26.54\% & 1.6110 & 26.01\% & 1.6242 & 27.71\% & 1.5392 & 22.84\% \\
20 & 1.7519 & 25.09\% & 1.7938& 22.18\%  & 1.7410 & 31.72\% & 1.7280 & 30.07\% & 1.7410 & 34.77\% & 1.6982 & 32.83\% & 1.7395 & 36.77\% & 1.6471 & 31.45\% \\
25 & 1.8270 & 30.45\% & 1.8018 & 27.49\% & 1.8307 & 38.12\% & 1.7474 & 31.53\% & 1.8307 & 41.72\% & 1.7730 & 38.68\% & 1.8221 & 43.27\% & 1.7830 & 42.3\% \\
\bottomrule
\end{tabular}}
\label{tab:robustness-rmse-rt}
\end{table*}

\begin{table*}[htbp]
\centering
\caption{MAE Comparison and Growth Rate under Different Noise Levels and Densities on TP Datasets}
\resizebox{\textwidth}{!}{
\begin{tabular}{c|cc|cc|cc|cc|cc|cc|cc|cc}
\toprule
\multirow{2}{*}{Noise (\%)} 
& \multicolumn{2}{c|}{QoSGNN@2.5} & \multicolumn{2}{c|}{QoSDiff@2.5} 
& \multicolumn{2}{c|}{QoSGNN@5} & \multicolumn{2}{c|}{QoSDiff@5}
& \multicolumn{2}{c|}{QoSGNN@7.5} & \multicolumn{2}{c|}{QoSDiff@7.5}
& \multicolumn{2}{c|}{QoSGNN@10} & \multicolumn{2}{c|}{QoSDiff@10} \\
& MAE & Degrad. & MAE & Degrad. & MAE & Degrad. & MAE & Degrad. 
& MAE & Degrad. & MAE & Degrad. & MAE & Degrad. & MAE & Degrad. \\
\midrule
0  & 17.2546 & --      & 16.0637 & --      & 13.4307 & --       & 13.1758 &  --      & 14.1491  & --      & 11.9951 & --      & 12.9670 & --       & 11.2097  & --   \\
5  & 21.0121 & 21.78\% & 18.9022 & 17.67\% & 17.6206 & 31.2\%   & 16.9981 & 29.01\%  & 16.6359  & 17.58\% & 14.0451 & 17.09\% & 16.1500 & 24.55\%  &13.8664 & 23.70\% \\
10 & 24.7343 & 43.35\% & 22.2836 & 38.72\% & 21.7548 & 61.98\%  & 20.7651 & 57.60\%  & 20.8886  & 47.63\% & 16.6696 & 38.97\% & 20.5445 & 58.44\%  & 16.4413 & 46.67\% \\
15 & 28.3657 & 64.40\% & 24.7252 & 53.92\% & 25.8768 & 92.67\%  & 24.5795 & 86.55\%  & 25.1842  & 77.99\% & 20.6496 & 72.15\% & 24.8885 & 91.94\%  & 20.6617 & 84.32\% \\
20 & 31.8865 & 84.80\% & 28.9468 & 80.20\% & 29.6307 & 120.6\%  & 26.6006 & 101.8\%  & 29.2754  & 106.9\% & 23.7479 & 97.98\% & 29.1007 & 124.4\%  & 24.2029 & 115.9\% \\
25 & 35.3414 & 104.8\% & 39.8672 & 116.9\% & 33.6044 & 150.2\%  & 30.1818 & 129.0\%  & 33.4169  & 136.1\% & 27.0621 & 125.6\% & 33.0010 & 153.5\%  & 27.3147 & 143.6\% \\
\bottomrule
\end{tabular}}
\label{tab:robustness-mae-tp}
\end{table*}

\begin{table*}[htbp]
\centering
\caption{RMSE Comparison and Growth Rate under Different Noise Levels and Densities on TP Datasets}
\resizebox{\textwidth}{!}{
\begin{tabular}{c|cc|cc|cc|cc|cc|cc|cc|cc}
\toprule
\multirow{2}{*}{Noise (\%)} 
& \multicolumn{2}{c|}{QoSGNN@2.5} & \multicolumn{2}{c|}{QoSDiff@2.5} 
& \multicolumn{2}{c|}{QoSGNN@5} & \multicolumn{2}{c|}{QoSDiff@5}
& \multicolumn{2}{c|}{QoSGNN@7.5} & \multicolumn{2}{c|}{QoSDiff@7.5}
& \multicolumn{2}{c|}{QoSGNN@10} & \multicolumn{2}{c|}{QoSDiff@10} \\
& RMSE & Degrad. & RMSE & Degrad. & RMSE & Degrad. & RMSE & Degrad. 
& RMSE & Degrad. & RMSE & Degrad. & RMSE & Degrad. & RMSE & Degrad. \\
\midrule
0  & 55.741 & --      & 54.115 & --      &  47.694 & --      & 46.329 & --      & 46.878 & --      & 43.303 & --      & 46.917 & --      &  41.086 & --   \\
5  & 68.865 & 23.55\% & 66.242 & 22.32\% &  62.026 & 35.74\% & 62.427 & 34.75\% & 61.283 & 30.73\% & 55.757 & 17.09\% & 60.526 & 29.00\% &  51.423 & 25.16\% \\
10 & 79.823 & 43.2\%  & 74.241 & 37.09\% &  74.888 & 63.89\% & 72.773 & 57.08\% & 74.431 & 58.77\% & 68.056 & 38.97\% & 74.369 & 58.51\% &  62.792 & 52.83\% \\
15 & 89.258 & 60.13\% & 83.323 & 53.86\% &  86.017 & 88.25\% & 85.656 & 84.89\% & 86.001 & 83.46\% & 73.612 & 72.15\% & 85.945 & 83.18\% &  70.257 & 71.29\% \\
20 & 97.605 & 75.11\% & 89.496 & 65.26\% &  94.762 & 107.3\% & 86.097 & 85.84\% & 95.908 & 104.5\% & 86.854 & 97.98\% & 96.390 & 105.4\% &  82.123 & 99.88\% \\
25 & 105.33 & 88.98\% & 94.240 & 74.02\% &  103.62 & 126.7\% & 95.334 & 105.8\% & 104.89 & 123.7\% & 93.674 & 125.6\% & 104.49 & 122.7\% &  86.597 & 110.77\% \\
\bottomrule
\end{tabular}}
\label{tab:robustness-rmse-tp}
\end{table*}

The experimental results reported in Tables~\ref{tab:robustness-mae-rt} through~\ref{tab:robustness-rmse-tp} offer a thorough assessment of the robustness of the proposed model under varying levels of synthetic noise and matrix sparsity. Several notable findings emerge from the empirical analysis:

\textbf{1. General Robustness Against Observational Noise.}
In the vast majority of tested scenarios across both RT and TP datasets, QoSDiff maintains a lower prediction error compared to the strong baseline, QoSGNN. While the baseline exhibits competitive performance in specific low-density settings (e.g., TP MAE at 2.5\% density with extreme noise), QoSDiff demonstrates superior overall stability, particularly in maintaining low RMSE values. This indicates that the diffusion-guided embedding learning effectively captures intrinsic data distributions, reducing the model's reliance on the precision of individual observed interactions.

\textbf{2. Controlled Performance Degradation.}
As the noise ratio increases from 5\% to 25\%, both models inevitably experience performance decay. However, the relative degradation (denoted as \textit{Degrad.}) of QoSDiff is generally more contained than that of QoSGNN. For instance, in the RT dataset, QoSDiff exhibits a flatter degradation curve in most density settings. This stability can be attributed to the Adversarial Attention-based Interaction Module (AAIM), which acts as a regularizer, filtering out incoherent interaction patterns introduced by the synthetic perturbations.

\textbf{3. Structural Independence in Sparse Environments.}
A key advantage of QoSDiff lies in its decoupling of embedding learning from explicit graph construction. Traditional GNN-based methods (like QoSGNN) are susceptible to error propagation when the underlying graph structure is corrupted by noisy edges, a problem that is exacerbated in sparse environments. By operating in a continuous latent space via the diffusion process, QoSDiff mitigates this structural sensitivity. The results confirm that even under significant noise injection, our framework retains its predictive capability, offering a reliable solution for real-world scenarios where service invocation logs may be sparse or unreliable.

\section{Related Work}
\label{sec:relatedwork}

In this section, we review the existing literature relevant to our work, categorized into three streams: traditional collaborative filtering, deep learning and graph-based approaches, and the emerging generative models for prediction tasks.

Early QoS prediction research was dominated by Collaborative Filtering (CF) techniques. Memory-based methods, such as UPCC \cite{UPCC}, IPCC \cite{IPCC}, and their hybrid UIPCC \cite{UIPCC}, predict missing QoS values by aggregating feedback from similar users or services based on Pearson Correlation Coefficients. While interpretable, these methods suffer severely from data sparsity and cold-start problems, as similarity computation becomes unreliable when the overlap between historical records is minimal.
To mitigate this issue, Matrix Factorization (MF) models like PMF \cite{PMF} and BiasMF \cite{BMF} were introduced to map users and services into low-dimensional latent spaces. Other models, such as AMF \cite{AMF} and NDMF \cite{NDMF}, utilize latent relationships to improve prediction accuracy. Although MF-based methods capture global data patterns more effectively than memory-based CF, they are inherently limited by their linear interaction modeling, failing to capture the complex, non-linear dependencies characteristic of large-scale Web service environments.

The success of deep learning in computer vision \cite{krizhevsky2012imagenet} inspired a wave of neural QoS prediction models. CSMF \cite{CSMF} and NCRL \cite{NCRL} leverage multi-layer perceptrons (MLPs) to learn non-linear feature representations. 
More recently, GNNs have become the state-of-the-art. Models such as GraphMF \cite{GraphMF}, QoSGNN \cite{QoSGNN}, and PMP \cite{he2024polarized} explicitly model user--service interactions as a bipartite graph, using message-passing mechanisms to aggregate neighbor information. 

To further address data sparsity, advanced GNN variants have been proposed: ISPAGNN \cite{ISPAGNN} introduces subgraph sampling to improve generalizability, while BGCL \cite{BGCL} employs contrastive learning to mitigate cold-start issues.

Despite these advancements, GNN-based methods face a critical bottleneck: \textit{dependence on explicit graph reliability}. 
Sampling-based methods risk information loss and bias, while standard message-passing models tend to propagate and amplify environmental noise through their layers, as demonstrated in our robustness experiments (Section \ref{sec:exp}-E). Furthermore, constructing explicit edges in massive, dynamic IoT environments is computationally prohibitive, highlighting a scalability gap in current structural learning paradigms.

Generative models, such as Variational Autoencoders (VAEs) \cite{kingma2013auto} and Generative Adversarial Networks (GANs) \cite{goodfellow2014generative}, have been explored to model the underlying data distribution for data imputation and recommendation tasks.
Recently, Denoising Diffusion Probabilistic Models (DDPMs) \cite{DDPM} have emerged as a powerful paradigm in recommender systems. Representative works like DiffRec \cite{wang2023diffusion} and DiffuRec \cite{li2023diffurec} formulate recommendations as a generative task, learning to reconstruct the user interaction matrix from Gaussian noise through a multi-step Markov chain process.

However, directly transplanting these standard diffusion frameworks to QoS prediction encounters non-trivial challenges.
First, standard DDPMs typically require a large number of diffusion timesteps ($T \gg 100$) for both training and inference. For real-time QoS prediction in large-scale service ecosystems, such iterative sampling is computationally prohibitive.
Moreover, existing methods often treat the interaction matrix as a grid-like image, employing CNN-based U-Nets \cite{unet} or simple MLPs for noise estimation. This design ignores the discrete and semantic nature of user/service identities and fails to capture the higher-order dependencies in the latent space.
Finally, the complex optimization of the full reverse process often suffers from convergence difficulties, limiting reproducibility in regression tasks.

Distinct from these approaches, \emph{QoSDiff} fundamentally rethinks the application of diffusion for QoS prediction. Instead of diffusing the raw interaction matrix, we propose a novel DELM that operates in the continuous latent space. Crucially, we reframe the embedding initialization as a single-step forward diffusion process, allowing us to perform efficient denoising in a single shot without the costly iterative chain.
Furthermore, unlike standard U-Net architectures, we design a specialized Attention-based Noise Predictor that naturally aligns with the non-Euclidean structure of user-service embeddings.
Rather than relying solely on the generative loss, we integrate this efficient diffusion mechanism with an AAIM, ensuring that the learned representations are not only robust to noise but also semantically discriminative for accurate QoS regression.

In summary, while the field has evolved from linear CF to complex GNN architectures, the trade-off between structural modeling capability and noise robustness remains a significant challenge. By bridging efficient latent diffusion with adversarial interaction learning, our work addresses these limitations, offering a scalable and resilient solution that outperforms existing paradigms in volatile service environments.

\section{Conclusion}
\label{sec:con}
In this paper, we presented \emph{QoSDiff}, a novel framework that fundamentally decouples robust representation learning from the constraints of explicit graph construction. By integrating a single-step diffusion-based embedding generator with an adversarial attention mechanism, our approach effectively circumvents the dual challenges of extreme data sparsity and environmental noise. Extensive evaluations on large-scale real-world datasets demonstrate that \emph{QoSDiff} significantly outperforms state-of-the-art baselines. These results provide compelling evidence that generative diffusion processes operating in the continuous latent space can capture intricate user--service dependencies more robustly than traditional message-passing paradigms, particularly in volatile service environments.

Looking forward, we envision two promising directions to extend this work:
\begin{itemize}
    \item \textbf{Structural Enhancement:} While the current self-attention mechanism effectively captures global dependencies, integrating more topology-aware architectures—such as hypergraph attention networks or hierarchical transformer blocks—could further refine the model's expressiveness. Future efforts will focus on leveraging these advanced structures to better characterize the complex, multi-modal correlations inherent in high-dimensional QoS data.
    
    \item \textbf{Temporal Dynamics:} The current framework focuses on static QoS prediction; however, real-world service quality is intrinsically dynamic. We plan to extend our diffusion paradigm to the temporal dimension by incorporating continuous time-series modeling. This enhancement will enable the framework to capture the evolution of user preferences and service variability, thereby significantly improving its applicability to real-time, dynamic service monitoring scenarios.
\end{itemize}






\section*{Acknowledgments}
This research was financially supported by Guangdong Basic and Applied Basic Research Foundation (No.2024A1515012468, 2024A1515011765), Special Projects in Key Fields of Guangdong Universities (No. 2022ZDZX1008), Guangdong Province Special Fund for Science and Technology (\"major special projects + task list\") Project (No. STKJ202209017), Guangdong Science and Technology Plan (No.STKJ2023012). Central Guiding Local Science and Technology Development Special Fund Project (No. STKJ2024083), Guangdong Provincial Characteristic Innovation Project of Regular Colleges and Universities (No.2025KTSCX035).

The authors would like to thank the computing support provided by the DELL Precision Workstation (3rd Gen), which was used to generate Figure \ref{fig1} in this paper.

\bibliographystyle{ieeetr}
\bibliography{references} 

@inproceedings{zeng2003quality,
  title={Quality driven web services composition},
  author={Zeng, Liangzhao and Benatallah, Boualem and Dumas, Marlon and Kalagnanam, Jayant and Sheng, Quan Z},
  booktitle={Proceedings of the 12th international conference on World Wide Web},
  pages={411--421},
  year={2003}
}

@article{UPCC,
  title={Collaborative filtering based recommendation system: A survey},
  author={Hameed, Mohd Abdul and Al Jadaan, Omar and Ramachandram, Sirandas},
  journal={International Journal on Computer Science and Engineering},
  volume={4},
  number={5},
  pages={859},
  year={2012},
  publisher={Engg Journals Publications}
}

@inproceedings{IPCC,
  title={Item-based collaborative filtering recommendation algorithms},
  author={Sarwar, Badrul and Karypis, George and Konstan, Joseph and Riedl, John},
  booktitle={Proceedings of the 10th international conference on World Wide Web},
  pages={285--295},
  year={2001}
}

@inproceedings{UIPCC,
  title={Collaborative reliability prediction of service-oriented systems},
  author={Zheng, Zibin and Lyu, Michael R},
  booktitle={Proceedings of the 32nd ACM/IEEE International Conference on Software Engineering-Volume 1},
  pages={35--44},
  year={2010}
}

@article{CSMF,
  title={Collaborative QoS prediction with context-sensitive matrix factorization},
  author={Wu, Hao and Yue, Kun and Li, Bo and Zhang, Binbin and Hsu, Ching-Hsien},
  journal={Future Generation Computer Systems},
  volume={82},
  pages={669--678},
  year={2018},
  publisher={Elsevier}
}

@article{AMF,
  title={Online QoS prediction for runtime service adaptation via adaptive matrix factorization},
  author={Zhu, Jieming and He, Pinjia and Zheng, Zibin and Lyu, Michael R},
  journal={IEEE Transactions on Parallel and Distributed Systems},
  volume={28},
  number={10},
  pages={2911--2924},
  year={2017},
  publisher={IEEE}
}

@article{NDMF,
  title={NDMF: Neighborhood-integrated deep matrix factorization for service QoS prediction},
  author={Zou, Guobing and Chen, Jin and He, Qiang and Li, Kuan-Ching and Zhang, Bofeng and Gan, Yanglan},
  journal={IEEE Transactions on Network and Service Management},
  volume={17},
  number={4},
  pages={2717--2730},
  year={2020},
  publisher={IEEE}
}

@article{NCF,
  title={Context-aware QoS prediction with neural collaborative filtering for Internet-of-Things services},
  author={Gao, Honghao and Xu, Yueshen and Yin, Yuyu and Zhang, Weipeng and Li, Rui and Wang, Xinheng},
  journal={IEEE Internet of Things Journal},
  volume={7},
  number={5},
  pages={4532--4542},
  year={2019},
  publisher={IEEE}
}

@article{WsDream,
  title={Investigating QoS of real-world web services},
  author={Zheng, Zibin and Zhang, Yilei and Lyu, Michael R},
  journal={IEEE transactions on services computing},
  volume={7},
  number={1},
  pages={32--39},
  year={2012},
  publisher={IEEE}
}

@article{PMF,
  title={Probabilistic matrix factorization},
  author={Mnih, Andriy and Salakhutdinov, Russ R},
  journal={Advances in neural information processing systems},
  volume={20},
  year={2007}
}

@inproceedings{BMF,
  title={Personalized QoS prediction for web services using latent factor models},
  author={Yu, Dongjin and Liu, Yu and Xu, Yueshen and Yin, Yuyu},
  booktitle={2014 IEEE international conference on services computing},
  pages={107--114},
  year={2014},
  organization={IEEE}
}

@article{NFMF,
  title={NFMF: neural fusion matrix factorisation for QoS prediction in service selection},
  author={Xu, Jianlong and Xiao, Lijun and Li, Yuhui and Huang, Mingwei and Zhuang, Zicong and Weng, Tien-Hsiung and Liang, Wei},
  journal={Connection Science},
  volume={33},
  number={3},
  pages={753--768},
  year={2021},
  publisher={Taylor \& Francis}
}

@article{NCRL,
  title={NCRL: Neighborhood-based collaborative residual learning for adaptive QoS prediction},
  author={Zou, Guobing and Wu, Shaogang and Hu, Shengxiang and Cao, Chenhong and Gan, Yanglan and Zhang, Bofeng and Chen, Yixin},
  journal={IEEE Transactions on Services Computing},
  volume={16},
  number={3},
  pages={2030--2043},
  year={2022},
  publisher={IEEE}
}

@article{DDPM,
  title={Denoising diffusion probabilistic models},
  author={Ho, Jonathan and Jain, Ajay and Abbeel, Pieter},
  journal={Advances in neural information processing systems},
  volume={33},
  pages={6840--6851},
  year={2020}
}

@inproceedings{GraphMF,
  title={GraphMF: QoS prediction for large scale blockchain service selection},
  author={Li, Yuhui and Xu, Jianlong and Liang, Wei},
  booktitle={2020 3rd International Conference on Smart BlockChain (SmartBlock)},
  pages={167--172},
  year={2020},
  organization={IEEE}
}

@article{QoSGNN,
  title={QoSGNN: Boosting QoS prediction performance with graph neural networks},
  author={Liu, Mingyi and Xu, Hanchuan and Sheng, Quan Z and Wang, Zhongjie},
  journal={IEEE Transactions on Services Computing},
  volume={17},
  number={2},
  pages={645--658},
  year={2023},
  publisher={IEEE}
}

@article{GNNs,
  title={The graph neural network model},
  author={Scarselli, Franco and Gori, Marco and Tsoi, Ah Chung and Hagenbuchner, Markus and Monfardini, Gabriele},
  journal={IEEE transactions on neural networks},
  volume={20},
  number={1},
  pages={61--80},
  year={2008},
  publisher={IEEE}
}

@article{BGCL,
  title={BGCL: Bi-subgraph network based on graph contrastive learning for cold-start QoS prediction},
  author={Zhu, Jiangyuan and Li, Bing and Wang, Jian and Li, Duantengchuan and Liu, Yongqiang and Zhang, Zhen},
  journal={Knowledge-Based Systems},
  volume={263},
  pages={110296},
  year={2023},
  publisher={Elsevier}
}

@inproceedings{ISPAGNN,
  title={Subgraph sampling for inductive sparse cloud services QoS prediction},
  author={Xu, Jianlong and Xia, Zhiyu and Li, Yuhui and Zeng, Yuxiang and Liu, Zhidan},
  booktitle={2022 IEEE 28th International Conference on Parallel and Distributed Systems (ICPADS)},
  pages={745--753},
  year={2023},
  organization={IEEE}
}

@article{peng2025energy,
  title={Energy-aware cloud manufacturing service selection and scheduling optimization},
  author={Peng, GaoXian and Wen, YiPing and Liu, JianXun and Kang, GuoSheng and Zhang, BiMing and Zhou, MinHao},
  journal={International Journal of Computer Integrated Manufacturing},
  volume={38},
  number={3},
  pages={309--334},
  year={2025},
  publisher={Taylor \& Francis}
}

@article{wu2024constraint,
  title={Constraint-aware and multi-objective optimization for micro-service composition in mobile edge computing},
  author={Wu, Jintao and Zhang, Jingyi and Zhang, Yiwen and Wen, Yiping},
  journal={Software: Practice and Experience},
  volume={54},
  number={9},
  pages={1596--1620},
  year={2024},
  publisher={Wiley Online Library}
}

@article{cao2024prkg,
  title={PRKG: pre-training representation and knowledge-graph-enhanced Web service recommendation for Mashup creation},
  author={Cao, Buqing and Peng, Mi and Xie, Ziming and Liu, Jianxun and Ye, Hongfan and Li, Bing and Fletcher, Kenneth K},
  journal={IEEE Transactions on Network and Service Management},
  volume={21},
  number={2},
  pages={1737--1749},
  year={2024},
  publisher={IEEE}
}

@ARTICLE{10660479,
author={Zhang, Heng and Huang, Shaoyuan and Xu, Mengwei and Guo, Deke and Wang, Xiaofei and Wang, Xin and Leung, Victor C. M. and Wang, Wenyu},
journal={ IEEE Transactions on Cloud Computing },
title={{ Large-Scale Measurements and Optimizations on Latency in Edge Clouds }},
year={2024},
volume={12},
number={04},
ISSN={2168-7161},
pages={1218-1231},
doi={10.1109/TCC.2024.3452094},
url = {https://doi.ieeecomputersociety.org/10.1109/TCC.2024.3452094},
publisher={IEEE Computer Society},
address={Los Alamitos, CA, USA},
month=oct}

@article{krizhevsky2012imagenet,
  title={Imagenet classification with deep convolutional neural networks},
  author={Krizhevsky, Alex and Sutskever, Ilya and Hinton, Geoffrey E},
  journal={Advances in neural information processing systems},
  volume={25},
  year={2012}
}

@inproceedings{unet,
  title={U-net: Convolutional networks for biomedical image segmentation},
  author={Ronneberger, Olaf and Fischer, Philipp and Brox, Thomas},
  booktitle={Medical image computing and computer-assisted intervention--MICCAI 2015: 18th international conference, Munich, Germany, October 5-9, 2015, proceedings, part III 18},
  pages={234--241},
  year={2015},
  organization={Springer}
}

@article{vaswani2017attention,
  title={Attention is all you need},
  author={Vaswani, Ashish and Shazeer, Noam and Parmar, Niki and Uszkoreit, Jakob and Jones, Llion and Gomez, Aidan N and Kaiser, {\L}ukasz and Polosukhin, Illia},
  journal={Advances in neural information processing systems},
  volume={30},
  year={2017}
}

@article{zou2025privacy,
  title={Privacy-Enhanced Federated Expanded Graph Learning for Secure QoS Prediction},
  author={Zou, Guobing and Yan, Zhi and Hu, Shengxiang and Gan, Yanglan and Zhang, Bofeng and Chen, Yixin},
  journal={IEEE Transactions on Services Computing},
  year={2025},
  publisher={IEEE}
}

@inproceedings{liu2025quality,
  title={Quality of Service Prediction via Large Language Models},
  author={Liu, Huiying and Zhang, Zekun and Sang, Lei and Wu, Qilin and Zhang, Yiwen},
  booktitle={2025 IEEE International Conference on Web Services (ICWS)},
  pages={87--95},
  year={2025},
  organization={IEEE}
}

@article{xing2025inv,
  title={Inv-Adapter: ID Customization Generation via Image Inversion and Lightweight Parameter Adapter},
  author={Xing, Peng and Wang, Ning and Ouyang, Jianbo and Li, Zechao},
  journal={IEEE Transactions on Pattern Analysis and Machine Intelligence},
  year={2025},
  publisher={IEEE}
}

@article{li2025dual,
  title={Dual-channel multiplex graph neural networks for recommendation},
  author={Li, Xiang and Fu, Chaofan and Zhao, Zhongying and Zheng, Guanjie and Huang, Chao and Yu, Yanwei and Dong, Junyu},
  journal={IEEE Transactions on Knowledge and Data Engineering},
  year={2025},
  publisher={IEEE}
}

@article{zhang2025survey,
  title={A survey on point-of-interest recommendation: Models, architectures, and security},
  author={Zhang, Qianru and Yang, Peng and Yu, Junliang and Wang, Haixin and He, Xingwei and Yiu, Siu-Ming and Yin, Hongzhi},
  journal={IEEE Transactions on Knowledge and Data Engineering},
  year={2025},
  publisher={IEEE}
}

@article{zeng2023gatcf,
  title={GATCF: Graph Attention Collaborative Filtering for Reliable Blockchain Services Selection in BaaS},
  author={Zeng, Yuxiang and Xu, Jianlong and Zhang, Zhuohua and Chen, Caiyi and Ling, Qianyu and Wang, Jialin},
  journal={Sensors},
  volume={23},
  number={15},
  pages={6775},
  year={2023},
  publisher={MDPI}
}

@article{hevapathige2025beyond,
  title={Beyond Fixed Depth: Adaptive Graph Neural Networks for Node Classification Under Varying Homophily},
  author={Hevapathige, Asela and Wijesinghe, Asiri and Zehmakan, Ahad N},
  journal={arXiv preprint arXiv:2511.06608},
  year={2025}
}

@article{wu2024robust,
  title={Robust QoS Prediction Based on Reputation Integrated Graph Convolution Network},
  author={Wu, Ziteng and Ding, Ding and Xiu, Yuting and Zhao, Yuekun and Hong, Jing},
  journal={IEEE Transactions on Services Computing},
  volume={17},
  number={03},
  pages={1154--1167},
  year={2024},
  publisher={IEEE Computer Society}
}

@article{he2024polarized,
  title={Polarized message-passing in graph neural networks},
  author={He, Tiantian and Liu, Yang and Ong, Yew-Soon and Wu, Xiaohu and Luo, Xin},
  journal={Artificial Intelligence},
  volume={331},
  pages={104129},
  year={2024},
  publisher={Elsevier}
}

@article{loshchilov2017decoupled,
  title={Decoupled weight decay regularization},
  author={Loshchilov, Ilya and Hutter, Frank},
  journal={arXiv preprint arXiv:1711.05101},
  year={2017}
}

@article{kingma2013auto,
  title={Auto-encoding variational bayes},
  author={Kingma, Diederik P and Welling, Max},
  journal={arXiv preprint arXiv:1312.6114},
  year={2013}
}

@article{goodfellow2014generative,
  title={Generative adversarial nets},
  author={Goodfellow, Ian J and Pouget-Abadie, Jean and Mirza, Mehdi and Xu, Bing and Warde-Farley, David and Ozair, Sherjil and Courville, Aaron and Bengio, Yoshua},
  journal={Advances in neural information processing systems},
  volume={27},
  year={2014}
}

@inproceedings{wang2023diffusion,
  title={Diffusion recommender model},
  author={Wang, Wenjie and Xu, Yiyan and Feng, Fuli and Lin, Xinyu and He, Xiangnan and Chua, Tat-Seng},
  booktitle={Proceedings of the 46th international ACM SIGIR conference on research and development in information retrieval},
  pages={832--841},
  year={2023}
}

@article{li2023diffurec,
  title={Diffurec: A diffusion model for sequential recommendation},
  author={Li, Zihao and Sun, Aixin and Li, Chenliang},
  journal={ACM Transactions on Information Systems},
  volume={42},
  number={3},
  pages={1--28},
  year={2023},
  publisher={ACM New York, NY}
}







\end{document}